\documentclass[letterpaper, 10 pt, conference]{ieeeconf}  %

\IEEEoverridecommandlockouts                              %

\let\labelindent\relax

\usepackage{booktabs}
\usepackage{bm}
\usepackage{multirow}
\usepackage{cite}

\usepackage{amsmath}
\usepackage{amsfonts}
\usepackage{graphicx}
\usepackage{adjustbox}
\usepackage{wrapfig}
\usepackage{xspace}
\usepackage{xcolor}
\usepackage{hyperref}
\hypersetup{
    colorlinks=true,
    linkcolor=orange,
    filecolor=magenta,      
    urlcolor=orange,
    citecolor=orange,
}
\usepackage[capitalise, nameinlink]{cleveref}
\crefname{lstlisting}{Listing}{Listings}
\usepackage{tabularx}
\usepackage{enumitem}
\usepackage{listings}
\usepackage{subcaption}
\usepackage{floatrow}
\newfloatcommand{capbtabbox}{table}[][\FBwidth]
\newcommand{\dataset}[0]{\textsc{PhysObjects}\xspace}
\newcommand{\website}[0]{https://iliad.stanford.edu/pg-vlm/}
\newcolumntype{C}[1]{>{\centering\let\newline\\\arraybackslash\hspace{0pt}}m{#1}}

\title{\LARGE \bf
Physically Grounded Vision-Language Models for Robotic Manipulation
}

\author{Jensen Gao$^1$, Bidipta Sarkar$^1$, Fei Xia$^{2}$, Ted Xiao$^2$, Jiajun Wu$^1$, \\ Brian Ichter$^2$, Anirudha Majumdar$^{2,3}$, Dorsa Sadigh$^{1,2}$%
\thanks{$^1$Stanford University, $^2$Google DeepMind, $^3$Princeton University. Contact: \texttt{jenseng@stanford.edu}.}%
}

\begin{document}

\maketitle
\thispagestyle{empty}
\pagestyle{empty}

\begin{abstract}
    Recent advances in vision-language models (VLMs) have led to improved performance on tasks such as visual question answering and image captioning.
    Consequently, these models are now well-positioned to reason about the physical world, particularly within domains such as robotic manipulation.
    However, current VLMs are limited in their understanding of the physical concepts (e.g., material, fragility) of common objects, which restricts their usefulness for robotic manipulation tasks that involve interaction and physical reasoning about such objects.
    To address this limitation, we propose \dataset, an object-centric dataset of 39.6K crowd-sourced and 417K automated physical concept annotations of common household objects.
    We demonstrate that fine-tuning a VLM on \dataset improves its understanding of physical object concepts, including generalization to held-out concepts, by capturing human priors of these concepts from visual appearance.
    We incorporate this physically grounded VLM in an interactive framework with a large language model-based robotic planner, and show improved planning performance on tasks that require reasoning about physical object concepts, compared to baselines that do not leverage physically grounded VLMs.
    We additionally illustrate the benefits of our physically grounded VLM on a real robot, where it improves task success rates.
    We release our dataset and provide further details and visualizations of our results at \url{\website}. \end{abstract}

\section{Introduction}
\label{sec:intro}

Large language models (LLMs) have shown great promise for converting language instructions into task plans for embodied agents \cite{huang2022language, ichter2022do}. The fundamental challenge in applying LLMs for this is grounding them to the physical world, through sensory input such as vision. Prior work has made progress towards grounding LLMs by using vision-language models (VLMs) to indicate the presence of objects in a scene, or to provide feedback about occurrences in a scene \cite{huang2022inner, chen2022nlmapsaycan, huang2023grounded, sharma2023semantic, wu2023tidybot}. However, vision could be used to further improve grounding by extracting more detailed scene information. For robotic manipulation, understanding physical concepts of objects, such as their material composition or their fragility, would help planners identify relevant objects to interact with, and affordances based on physical or safety constraints. For example, if a human wants a robot to get a cup of water, the robot should be able to determine if a cup already has water or something else in it. Also, the robot should handle the cup with greater caution if it is more fragile.

How can we use vision to reason about physical object concepts? Prior work has studied this problem using more traditional vision techniques, such as self-supervised learning on object interaction data. However, object interaction data can be challenging to collect when scaling up beyond a small set of objects in well-defined settings. While precise estimation of physical properties may sometimes be impossible without interaction data, humans can use their visual perception to reason at a high level about physical concepts without object interactions. For example, humans can reason that a glass cup is more fragile than a plastic bottle, and that it would be easier to use a bowl to hold water than a shallow plate. This reasoning is often based on prior semantic knowledge of visually similar objects, and can be done from static visual appearance alone.

Similarly, VLMs pre-trained using large-scale data have demonstrated broad visual reasoning abilities and generalization \cite{alayrac2022flamingo, chen2023pali, li2023blip, dai2023instructblip, liu2024prismer, driess2023palme}, and thus have the potential to physically reason about objects in a similar fashion as humans. Therefore, we propose to leverage VLMs as a scalable way of providing the kind of high-level physical reasoning that humans use to interact with the world, which can benefit a robotic planner, without the need for interaction data. The general and flexible nature of VLMs also removes the need to use separate task-specific vision models for physical reasoning. VLMs have already been commonly incorporated into robotic planning systems \cite{huang2022inner, chen2022nlmapsaycan, huang2023grounded, sharma2023semantic, driess2023palme, wu2023tidybot}, making them a natural solution for endowing physical reasoning into robotic planning.

\begin{figure*}[t]
    \vspace{5pt}
    \includegraphics[width=\linewidth]{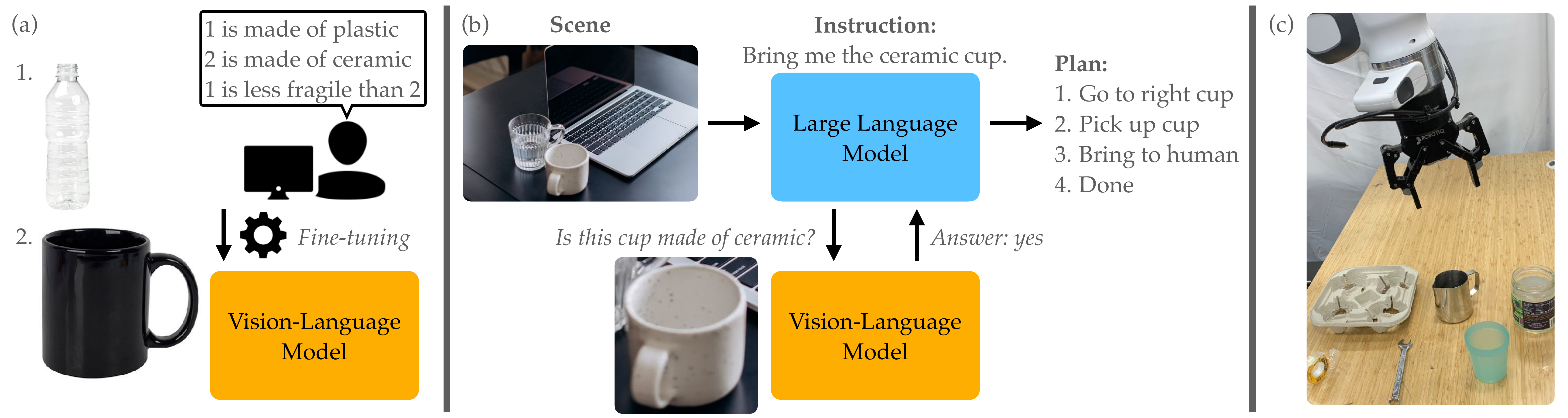}
    \centering
    \caption{(a) We collect physical concept annotations of common household objects for fine-tuning VLMs. (b) We use the fine-tuned VLM in an LLM-based robotic planning framework, where the LLM queries the VLM about physical concepts of objects in the scene, before producing a plan. (c) We evaluate LLM-generated plans on a real Franka Emika Panda robot. \vspace{-25pt}}%
    \label{fig:teaser}
\end{figure*}

However, while modern VLMs have improved significantly on tasks such as visual question answering (VQA), and there has been evidence of their potential for object-centric physical reasoning \cite{Peng2023OpenScene}, we show in this work that their out-of-the-box performance for this still leaves much to be desired. Although VLMs have been trained on broad internet-scale data, this data does not contain many examples of object-centric physical reasoning. This motivates incorporating a greater variety and amount of such data when training VLMs. Unfortunately, prior visual datasets for physical reasoning are not well-suited for understanding common real-world objects, which is desirable for robotics. To address this, we propose \dataset, an object-centric dataset with human physical concept annotations of common household objects. Our annotations include categorical labels (e.g., object X is made of plastic) and preference pairs (e.g., object X is heavier than object Y).

Our main contributions are \dataset, a dataset of 39.6K crowd-sourced and 417K automated physical concept annotations of real household objects, and demonstrating that using it to fine-tune a VLM significantly improves physical reasoning. We show that our physically grounded VLM achieves improved test accuracy on our dataset, including on held-out physical concepts. Furthermore, to illustrate the utility of improved physical reasoning for robotics, we incorporate our physically grounded VLM with an LLM-based robotic planner, where the LLM queries the VLM about physical concepts of objects in its scene. Our system achieves improved planning performance on tasks that require physical reasoning, compared to baselines that do not use physically grounded VLMs. Finally, we demonstrate the benefits of our physically grounded VLM for planning with a real robot, where its usage improves task success rates.
\vspace{-4pt}
\section{Related Work}
\label{sec:related_work}

We review prior work on physical reasoning, object attribute datasets, VLMs, using LLMs for robotic planning, and using LLMs and VLMs together in an interactive system.

\noindent \textbf{Physical Reasoning.}
Prior works have studied estimating physical object properties from vision by learning from interaction data \cite{wu2015galileo, wu2016physics, li2020visual}. Other works focus on learning representations that capture physical concepts, rather than direct estimation \cite{janner2019reasoning, DensePhysNet}. Unlike these works, we use pre-trained VLMs and human annotations as a more scalable alternative to learning from interaction. Mind's Eye investigates physical reasoning using LLMs \cite{liu2023minds}, but relies on grounding using a simulator, which would be difficult to scale to the real world. VEC investigates physical reasoning with LLMs and VLMs \cite{li2023can}, but reasons from text descriptions, while we reason from real images. OpenScene uses CLIP \cite{Radford2021LearningTV} to identify objects in scenes using properties such as material and fragility, but these results are only qualitative in nature \cite{Peng2023OpenScene}. In our work, we propose \dataset to better quantify and improve object-centric physical reasoning, and leverage this reasoning for robotic manipulation.

\noindent \textbf{Object Attribute Datasets.}
There have been prior visual object attribute datasets with concepts included in \dataset, such as material and transparency \cite{patterson2016coco, krishna2017visual, pham2021learning, Ramanathan_2023_CVPR}. However, they focus more on visual attributes such as color, while we focus on physical concepts. Physics 101 provides a dataset of object interaction videos and property measurements \cite{wu2016physics}, but \dataset includes a greater variety of objects that are more relevant for household robotics.

\noindent \textbf{Vision-Language Models.}
VLMs have made large improvements on multi-modal tasks such as VQA, by leveraging internet-scale image and text data \cite{alayrac2022flamingo, chen2023pali, li2023blip, liu2024prismer}. In our experiments, we use InstructBLIP \cite{dai2023instructblip} as our base VLM for fine-tuning and comparison, as it was the state-of-the-art open-source VLM at the time of our experiments. PaLM-E has shown strong performance on general visual-language tasks and robotic planning \cite{driess2023palme}, but there has not been focused evaluation of it for physical reasoning. SuccessVQA fine-tunes VLMs on human data for success detection by treating it as a VQA task, and achieves better generalization than models designed specifically for success detection \cite{du2023visionlanguage}. We similarly fine-tune VLMs on human data for physical reasoning by casting it as a VQA problem, to benefit from the generalization abilities and versatility of VLMs.

\noindent \textbf{LLMs for Robotic Planning.}
Many recent works have used LLMs as robotic planners. SayCan uses visual value functions to provide affordances for grounding \cite{ichter2022do}, but does not benefit from VLMs. Follow-up works have used VLMs for grounding LLM planners through object detection, or providing feedback about what has happened (e.g., success detection) \cite{huang2022inner, chen2022nlmapsaycan, huang2023grounded, sharma2023semantic, wu2023tidybot}. Our work focuses on expanding the use of VLMs for grounding through physical reasoning, to let LLM-based planners perform tasks that require a deeper physical understanding of the world.

\noindent \textbf{LLM/VLM Interaction.}
Our planning evaluation falls in the framework of Socratic Models \cite{zeng2023socratic}, where large models interact with each other through text to perform tasks such as VQA \cite{yang2022empirical, shao2023prompting} and image captioning \cite{zhu2023chatgpt}. Most similar to our evaluation is Matcha, where an LLM receives a task instruction, obtains object-centric feedback from its environment, and uses this for task planning \cite{zhao2023chat}. However, this work does not focus on visual feedback, as their evaluation is in a simulated environment where physical concepts are not visually observable. In contrast, we focus on physical reasoning from vision in real-world scenes.
\vspace{-4pt}
\section{\dataset Dataset}
\label{sec:dataset}

To benchmark and improve VLMs for object-centric physical reasoning, we propose \dataset, a dataset of 39.6K crowd-sourced and 417K automated physical concept annotations for images of real household objects.

\noindent{\textbf{Image Source.}}
We use the publicly released challenge version of the EgoObjects dataset \cite{meta2022ego} as our image source. To our knowledge, this was the largest object-centric dataset of real images that was publicly released when constructing \dataset. The dataset consists of frames from egocentric videos in realistic household settings, which makes it particularly relevant for household robotics. It includes 117,424 images, 225,466 object bounding boxes with corresponding category labels from 277 object categories, and 4,203 object instance IDs. \dataset consists of physical concept annotations for a large subset of this image data.
\footnote{We publicly release our dataset on our \href{\website}{website}. Because the EgoObjects license does not permit incorporating it into another dataset, we release our annotations separately from the image data.}

We construct random training, validation, and test sets based on object instance IDs. We split the dataset per object category to ensure each object category is represented in each set when possible. Our training, validation, and test sets consist of 73.0\%, 14.8\%, and 12.2\% of objects, respectively.

\begin{table}[ht]
    \vspace{-3pt}
    \begin{adjustbox}{width=\textwidth}
    \begin{tabular}{ll}
        \toprule
        \textbf{Concept} & \textbf{Description} \\
        \midrule
        Mass                              & how heavy an object is                                 \\
        Fragility                         & how easily an object can be broken/damaged             \\
        Deformability                     & how easily an object can change shape without breaking \\
        Material                          & what an object is primarily made of                    \\
        Transparency                      & how much can be seen through an object                 \\
        Contents                          & what is inside a container                             \\
        Can Contain Liquid                & if a container can be used to easily carry liquid      \\
        Is Sealed                         & if a container will not spill if rotated               \\
        \midrule
        Density (\emph{held-out})         & how much mass per unit of volume of an object          \\
        Liquid Capacity (\emph{held-out}) & how much liquid a container can contain                \\
        \bottomrule
    \end{tabular}
    \end{adjustbox}
    \caption{Our physical concepts and brief descriptions \vspace{-5pt}}%
    \label{table:phys_char}
\end{table}

\noindent{\textbf{Physical Concepts.}}
We collect annotations for eight main physical concepts and two additional concepts reserved for held-out evaluation. We select concepts based on prior work and what we believe to be useful for robotic manipulation, but do not consider all such concepts. For example, we do not include \emph{friction} because this can be challenging to estimate without interaction, and we do not include \textit{volume} because this requires geometric reasoning, which we do not focus on.

Of our main concepts, three are continuous-valued and applicable to all objects: \emph{mass}, \emph{fragility}, and \emph{deformability}. Two are also applicable to all objects, but are categorical: \emph{material} and \emph{transparency}. \emph{Transparency} could be considered continuous, but we use discrete values of \emph{transparent}, \emph{translucent}, and \emph{opaque}. The other three are categorical and applicable only to container objects: \emph{contents}, \emph{can contain liquid}, and \emph{is sealed}. We define which object categories are containers, resulting in 956 container object instances. 

Our two held-out concepts are \emph{density}, which is continuous and applicable to all objects, and \emph{liquid capacity}, which is continuous and applicable only to containers. We only collect test data for these held-out concepts. We list all concepts and their brief descriptions in \cref{table:phys_char}.

For categorical concepts, we define a set of labels for each concept. Annotations consist of a label specified for a given object and concept. For the concepts \emph{material} and \emph{contents}, when crowd-sourcing, we allow for open-ended labels if none of the pre-defined labels are applicable.

For continuous concepts, annotations are preference pairs, where given two objects, an annotation indicates that either one object has a higher level of a concept, the objects have roughly \emph{equal} levels, or the relationship is \emph{unclear}. We use preferences because it is generally more intuitive for humans to provide comparisons than continuous values \cite{sadigh2017active, NIPS2017_d5e2c0ad}. This is especially true when annotating static images with physical concepts, where it is difficult to specify precise grounded values. For example, it would be difficult to specify the \emph{deformability} of a sponge as a value out of 10. Comparisons have also been used to evaluate LLMs and VLMs for physical reasoning in prior work \cite{li2023can}. Therefore, the kind of grounding studied in \dataset for continuous concepts is only relational in nature.

\noindent{\textbf{Automatic Annotations.}}
Before crowd-sourcing, we first attempt to automate as many annotations as possible, so that crowd-workers only annotate examples that cannot be easily automated. For categorical concepts, we assign concept values to some of the defined object categories in EgoObjects, such that all objects in a category are labeled with that value. For continuous concepts, we define \emph{high} and \emph{low} tiers for each concept, such that all objects from a \emph{high} tier category have a higher level of that concept than all objects from a \emph{low} tier category. Then, we automate preference annotations for all object pairs between the two tiers.

\begin{figure}[ht]
    \includegraphics[width=\linewidth]{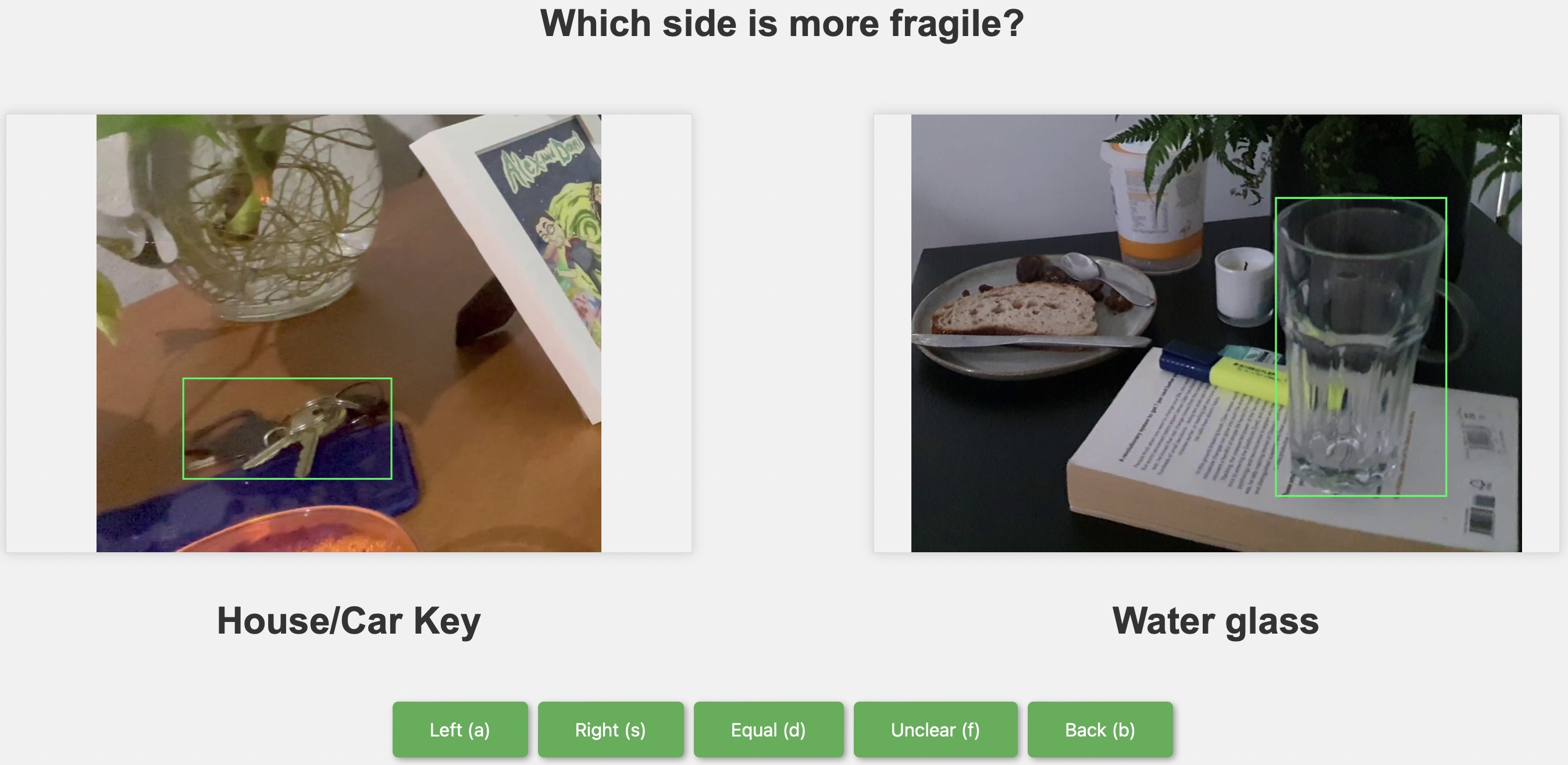}
    \caption{\centering Annotation UI for \emph{fragility}. Here, the label is \emph{right}, i.e., the \emph{water glass} is more fragile than the \emph{house/car key}. \vspace{-5pt}}%
    \label{fig:interface}
\end{figure}

\begin{table*}[t]
    \vspace{7pt}
    \centering
    \begin{adjustbox}{width=\textwidth}
    \begin{tabular}{lccccc}
        \toprule
                            & Most Common & Text Only       & InstructBLIP & Single Concept FT (ours) & PG-InstructBLIP (ours) \\
        \midrule
        Mass                & 42.2        & 73.3            & 62.2         & \textbf{80.0}            & \textbf{80.0}          \\
        Fragility           & 64.9        & 64.9            & 78.4         & 91.2                     & \textbf{94.6}          \\
        Deformability       & 46.5        & 62.8            & 67.4         & \textbf{95.3}            & 93.0                   \\
        Material            & 37.1        & 73.9            & 67.1         & 83.7                     & \textbf{84.6}          \\
        Transparency        & 77.6        & 82.2            & 85.8         & 89.4                     & \textbf{90.1}          \\
        Contents            & 39.5        & 50.9            & 35.1         & 81.6                     & \textbf{83.3}          \\
        Can Contain Liquid  & 56.3        & {\textbf{92.2}} & 59.4         & 84.4                     & 87.5                   \\
        Is Sealed           & 80.6        & 80.6            & 74.2         & 80.6                     & \textbf{87.1}          \\
        \midrule
        Average             & 55.6        & 72.6            & 66.2         & 85.8                     & \textbf{87.5}          \\       
        \bottomrule
    \end{tabular}
    \end{adjustbox}
    \caption{Test accuracy for main concepts on crowd-sourced \dataset \vspace{-15pt}}%
    \label{table:crowd_results}
\end{table*}

\noindent{\textbf{Crowd-Sourcing Annotations.}}
We obtain additional annotations via crowd-sourcing, using 573 crowd-workers on the Prolific platform. Crowd-workers use a web-based user interface (example for \emph{fragility} shown in \cref{fig:interface}) where they are presented with object bounding boxes in the context of their overall image, and provide annotations using on-screen buttons or their keyboard.
For categorical concepts, we collect annotations for the majority of objects that were not automatically annotated. For continuous concepts, because it is impractical to annotate every pair of objects in the dataset, we randomly sample pairs to annotate. We enforce that 20\% of the sampled pairs are between objects of the same category, to prioritize understanding differences between objects of the same category. We collect annotations from three crowd-workers for each example.
To promote high-quality data, we include attention checks as 10\% of provided examples, which have known labels, and only keep data from annotators that achieve 80\% accuracy on these.

\noindent{\textbf{Dataset Statistics.}}
We crowd-source 39.6K annotations for 13.2K examples, and automate annotations for 417K additional examples. For crowd-sourced annotations, 93.7\% of examples have at least 2/3 annotator label agreement, and 58.1\% have unanimous agreement.
\vspace{-3pt}
\section{Physically Grounding Vision-Language Models}
\label{sec:approach}

\noindent{\textbf{Fine-Tuning VLMs.}}
We work with the FlanT5-XXL \cite{wei2022finetuned} version of InstructBLIP \cite{dai2023instructblip}. InstructBLIP takes as input a single RGB image and text prompt, and predicts text as output. In our setup, we choose the model inputs to be a single bounding box of an object, and a question text prompt corresponding to each concept.

\noindent{\textbf{Learning From Preferences.}} Learning for categorical concepts amounts to maximum likelihood of annotated labels. However, it is not as straightforward to train a VLM on preferences for continuous concepts, because preference learning requires a continuous score. To do this with VLMs, which naturally have discrete text outputs, we prompt the VLM with questions that can be answered with \emph{yes} or \emph{no} for continuous concepts. Then, we extract the following score function:
\begin{equation*}
s(o, c) = \frac{p(\text{yes} \mid o, c)}{p(\text{no} \mid o, c)} \label{pref_score}
\end{equation*}
where $o$ is an object bounding box image, $c$ is a concept, and $p(\cdot| o, c)$ is the likelihood under the VLM of text, conditioned on the object image and concept. We use this as our score function because it can take any non-negative value, and $\log s(o, c)$ has the intuitive interpretation as the difference of log-likelihoods between \emph{yes} and \emph{no}.\footnote{We experimented with other choices of score functions, and found that while all performed similarly with respect to test accuracy on \dataset, we found this score function to produce the most interpretable range of likelihoods for different responses, which we hypothesize to be beneficial for downstream planning.} We then use the Bradley-Terry model \cite{bradley1952rank} to estimate the probability of a human indicating that object $o_1$ has a higher value than object $o_2$ for concept $c$ as:
\begin{equation*}
P(o_1 > o_2 \mid c) = \frac{s(o_1, c)}{s(o_1, c) + s(o_2, c)}. \label{pref_prob}
\end{equation*}
We assume a dataset $\mathcal{D}$ of preference annotations $(o_1, o_2, c, y)$, where $y \in \{[1, 0], [0, 1], [0.5, 0.5]\}$ corresponds to if $o_1$ is preferred, $o_2$ is preferred, or if they are indicated to be equal. We then fine-tune the VLM by minimizing the following objective:
\begin{align*}
    \mathcal{L}(\mathcal{D}) = &-\mathbb{E}_{(o_1, o_2, c, y) \sim \mathcal{D}}[y_1 \log P(o_1 > o_2 \mid c) \\
    &+ y_2 \log (1 - P(o_1 > o_2 \mid c)].
\end{align*}
In practice, this is the binary cross-entropy objective where the logits for each object image $o$ is the difference of log-likelihoods $\log s(o, c) = \log p(\text{yes} \mid o, c) - \log p(\text{no} \mid o, c)$.
\section{Experimental Results}
\label{sec:results}

We evaluate VLMs for physical reasoning using 1) test accuracy on \dataset, 2) planning accuracy on real scenes for physical reasoning tasks, and 3) task success rate on a real robot.

\subsection{Dataset Evaluation}
We refer to InstructBLIP fine-tuned on all main concepts in \dataset as Physically Grounded InstructBLIP, or PG-InstructBLIP. \footnote{We release the model weights for PG-InstructBLIP on our \href{\website}{website}.} We focus our evaluation on crowd-sourced examples, because as described in \cref{sec:dataset}, these were collected with the intent for their labels to not be discernible from object category information alone, and thus they are generally more challenging. We report test accuracy on these examples in \cref{table:crowd_results}. Our baselines include \emph{Most Common}, where the most common label in the training data is predicted, \emph{Text Only}, where an LLM makes predictions using in-context examples from \dataset, but using object category labels instead of images, and InstructBLIP. We also compare to versions of InstructBLIP fine-tuned on single concept data. We find that PG-InstructBLIP outperforms InstructBLIP on all concepts, with the largest improvement on \emph{contents}, which InstructBLIP has the most difficulty with. We also find that PG-InstructBLIP performs slightly better than the single concept models, suggesting possible positive transfer from using a single general-purpose model compared to separate task-specific models, although we acknowledge the improvement here is not extremely significant. PG-InstructBLIP also generally outperforms \emph{Most Common} and \emph{Text Only}, suggesting that our evaluation benefits from reasoning beyond dataset statistics, and from using vision.

\begin{table}[t]
    \small
    \vspace{7pt}
    \begin{tabular}[b]{p{3cm}C{1.5cm}C{2.6cm}}
        \toprule
                        & InstructBLIP  & PG-InstructBLIP (ours) \\
        \midrule
        Density         & 54.2          & \textbf{70.3}          \\
        Liquid Capacity & 65.4          & \textbf{73.0}          \\
        \midrule
        Average         & 59.8          & \textbf{71.7}          \\
        \bottomrule
    \end{tabular}
    \vspace{-3pt}
    \caption{\centering Test accuracy for held-out concepts on crowd-sourced \dataset \vspace{-3pt}}%
    \label{table:heldout}
\end{table}

\noindent \textbf{Generalization Results.} We additionally evaluate both InstructBLIP and PG-InstructBLIP on test data for our held-out concepts, which we report in \cref{table:heldout}. We find that PG-InstructBLIP improves upon InstructBLIP by \textbf{11.9\%}, despite having never seen these evaluated concepts nor object instances during fine-tuning. We believe this suggests that fine-tuning VLMs can offer possible generalization benefits to concepts that are related to those seen during fine-tuning.

\begin{table}[ht]
    \small
    \vspace{-5pt}
    \begin{tabular}[b]{p{3cm}C{1.5cm}C{2.6cm}}
        \toprule
                            & InstructBLIP & PG-InstructBLIP (ours) \\
        \midrule
        Mass                & 55.6         & \textbf{82.2}          \\
        Fragility           & 70.3         & \textbf{83.8}          \\
        Deformability       & 76.7         & \textbf{88.4}          \\
        Material            & 67.7         & \textbf{83.4}          \\
        Transparency        & 81.5         & \textbf{83.8}          \\
        Contents            & 32.5         & \textbf{81.6}          \\
        Can Contain Liquid  & 56.3         & \textbf{89.1}          \\
        Is Sealed           & 71.0         & \textbf{80.6}          \\
        \midrule
        Average             & 64.0         & \textbf{84.1}          \\
        \bottomrule
    \end{tabular}
    \vspace{-3pt}
    \caption{\centering Test accuracy for main concepts with paraphrased prompts \vspace{-3pt}}%
    \label{table:paraphrase}
\end{table}

\vspace{-5pt}
\noindent In \cref{table:paraphrase}, we report results for main concepts on unseen paraphrased question prompts. We find that PG-InstructBLIP still outperforms InstructBLIP, with limited degradation from the original prompts, suggesting robustness to question variety from using a large pre-trained VLM.

\begin{wrapfigure}{l}{0.5\linewidth}
    \vspace{-15pt}
    \includegraphics[width=\linewidth]{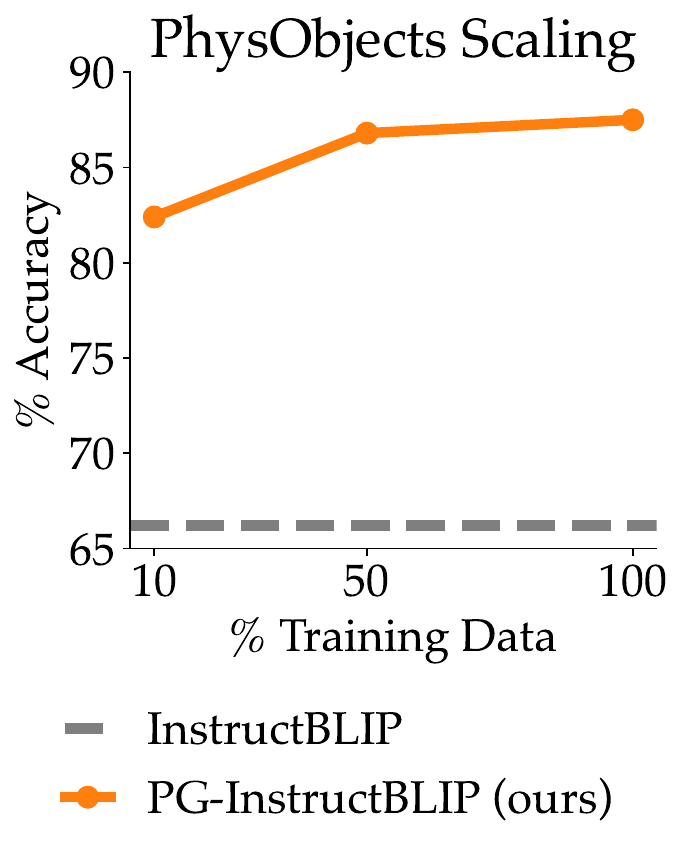}%
    \vspace{-10pt}
    \caption{\centering Performance scaling with dataset size \vspace{-10pt}}%
    \label{fig:dataset_scale}
\end{wrapfigure}

\vspace{5pt}
\noindent \textbf{Dataset Scaling.} In \cref{fig:dataset_scale}, we illustrate how average performance scales with dataset size, by fine-tuning on different fractions of data from \dataset. Performance scales positively, but the models still benefit significantly from only 10\% of our dataset, suggesting that the physical reasoning of VLMs can be improved with relatively small amounts of annotated data.

\vspace{2pt}
\noindent{\textbf{Additional Results.}}
We include additional results in our Appendix (found on our \href{\website}{website}). These include showing that PG-InstructBLIP has limited degradation on general VQA benchmarks compared to InstructBLIP, suggesting that existing systems using VLMs can benefit from \dataset for physical reasoning, without sacrificing other reasoning abilities. We also include results using different question prompts, using a smaller version of InstructBLIP, evaluating on automatically annotated data, transfer to held-out concepts, and ablations on our fine-tuning process.

\vspace{-3pt}
\subsection{Real Scene Planning Evaluation}
\begin{wrapfigure}{r}{0.5\linewidth}
    \includegraphics[width=\linewidth]{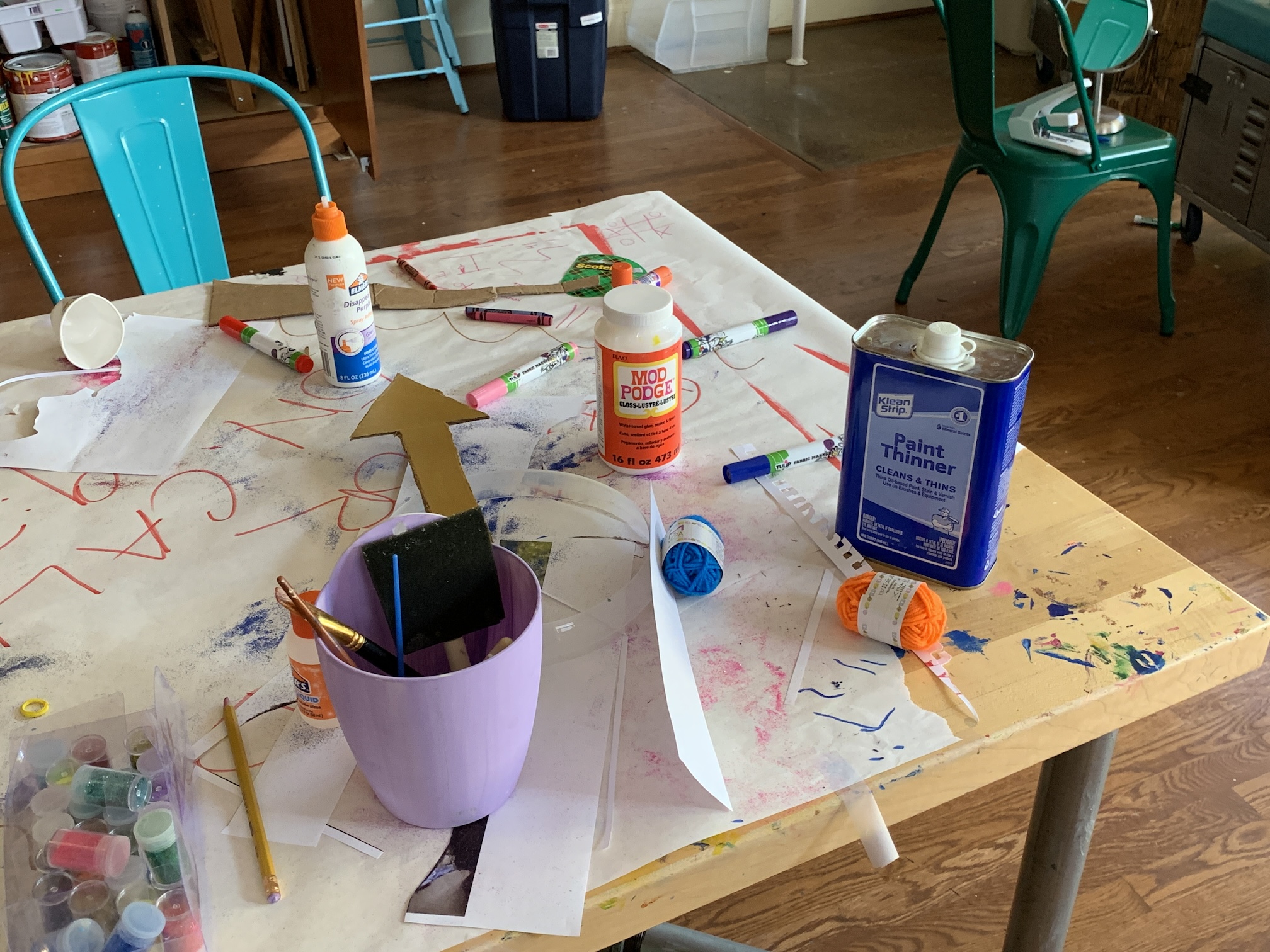}%
    \caption{\centering Example scene in our planning evaluation \vspace{-10pt}}%
    \label{fig:scene_ex}
\end{wrapfigure}

Next, we evaluate the efficacy of PG-InstructBLIP for robotic planning on unseen images of real scenes. We provide an example scene in \cref{fig:scene_ex}. We evaluate on tasks with language instructions, and assume a library of primitive robotic operations with language descriptions.

\noindent{\textbf{Planning Framework.}} The LLM used in our planning framework is GPT-4 \cite{openai2023gpt4}. It is first given object detections in the scene, a list of primitives, and the task instruction, and then asks a VLM questions about objects in the scene. There are no constraints on the questions. Afterwards, the LLM either indicates the task is not possible, or produces a plan consisting of primitives to execute.

\begin{table}[ht]
    \vspace{-6pt}
    \begin{adjustbox}{width=\textwidth}
    \begin{tabular}{p{3.25cm}C{1cm}C{1.75cm}C{2.6cm}}
        \toprule
        Task Category    & No VLM & InstructBLIP & PG-InstructBLIP (ours) \\ 
        \midrule
        Single Concept   & 36.8   & 68.4         & \textbf{84.1}          \\
        Multi-Concept    & 27.8   & 27.8         & \textbf{94.4}          \\
        Common Knowledge & 35.7   & 78.6         & \textbf{85.7}          \\
        \midrule
        Overall          & 33.3   & 56.9         & \textbf{88.2}          \\
        \bottomrule
    \end{tabular}
    \end{adjustbox}
    \caption{Task plan accuracy on 51 real scenarios \vspace{-5pt}}%
    \label{table:plan_results}
\end{table}

\noindent{\textbf{Results.}}
We report task planning accuracy using InstructBLIP and PG-InstructBLIP in \cref{table:plan_results}. We also compare to a planner that does not use VLM interaction for grounding. We evaluate on 51 task scenarios across 8 scenes, using a non-author human to evaluate task plans. We divide our task scenarios into three categories. \emph{Single Concept} requires identifying objects using one physical concept, e.g., finding the heaviest object. \emph{Multi-Concept} requires reasoning about multiple physical concepts, e.g., asking for a metal container that can hold water. This may include concepts outside of \dataset. \emph{Common Knowledge} requires additional reasoning about common knowledge of objects, e.g., understanding the label of a container. While our tasks focus on physical concepts in \dataset, the LLM can ask questions about other concepts that may also be useful, particularly for \emph{Common Knowledge} tasks.

PG-InstructBLIP outperforms InstructBLIP on all task categories, especially \emph{Multi-Concept}. It does slightly better on \emph{Common Knowledge}, suggesting that it can reason about non-\dataset concepts at a similar level as InstructBLIP. Using no VLM performs substantially worse than using VLM interaction, indicating that our tasks require additional grounding beyond object detection.
We provide further details of results on our \href{\website}{website}.

\subsection{Real Robot Evaluation}
Lastly, we evaluate plans on real scenes using a Franka Emika Panda robot. We use a similar planner as in the previous section, but with different prompts and primitives. We assume a library of primitives for pick-and-place tasks. We evaluate on two scenes, with five tasks per scene, which we provide in \cref{table:robot_ex}. We report success rates using InstructBLIP and PG-InstructBLIP in \cref{table:real_results}. We ensure the primitives execute successfully, so our success rates only reflect plan quality.

\begin{table}[ht]
    \small
    \centering
    \begin{tabular}{lp{4.25cm}}
        \toprule
        \textbf{Scene Image}                                   &  \textbf{Task Instructions} \\
        \midrule
        \raisebox{-\totalheight}{\includegraphics[width=0.4\textwidth]{figures/robot_scenes/robot_scene1_ex.JPG}} &
        \begin{enumerate}[leftmargin=*]
            \item Move all objects that are not plastic to the side.
            \item Find a container that has metals. Move all metal objects into that container.
            \item Move all containers that can be used to carry water to the side.
            \item Put the two objects with the least mass into the least deformable container.
            \item Move the most fragile object to the side.
        \end{enumerate} \\
        \raisebox{-\totalheight}{\includegraphics[width=0.4\textwidth]{figures/robot_scenes/robot_scene2_ex.JPG}} &
        \begin{enumerate}[leftmargin=*]
            \item Put all containers that can hold water to the side.
            \item Put all objects that are not plastic to the side.
            \item Put all objects that are translucent to the side.
            \item Put the three heaviest objects to the side.
            \item Put a plastic object that is not a container into a plastic container. Choose the container that you are most certain is plastic.
        \end{enumerate} \\
        \bottomrule
    \end{tabular}
    \caption{\centering Scene images and task instructions for our real robot evaluation \vspace{-5pt}}%
    \label{table:robot_ex}
\end{table}

\begin{figure}[ht]
    \vspace{-7pt}
    \begin{subfigure}[t]{0.575\columnwidth}
        \includegraphics[height=3.65cm]{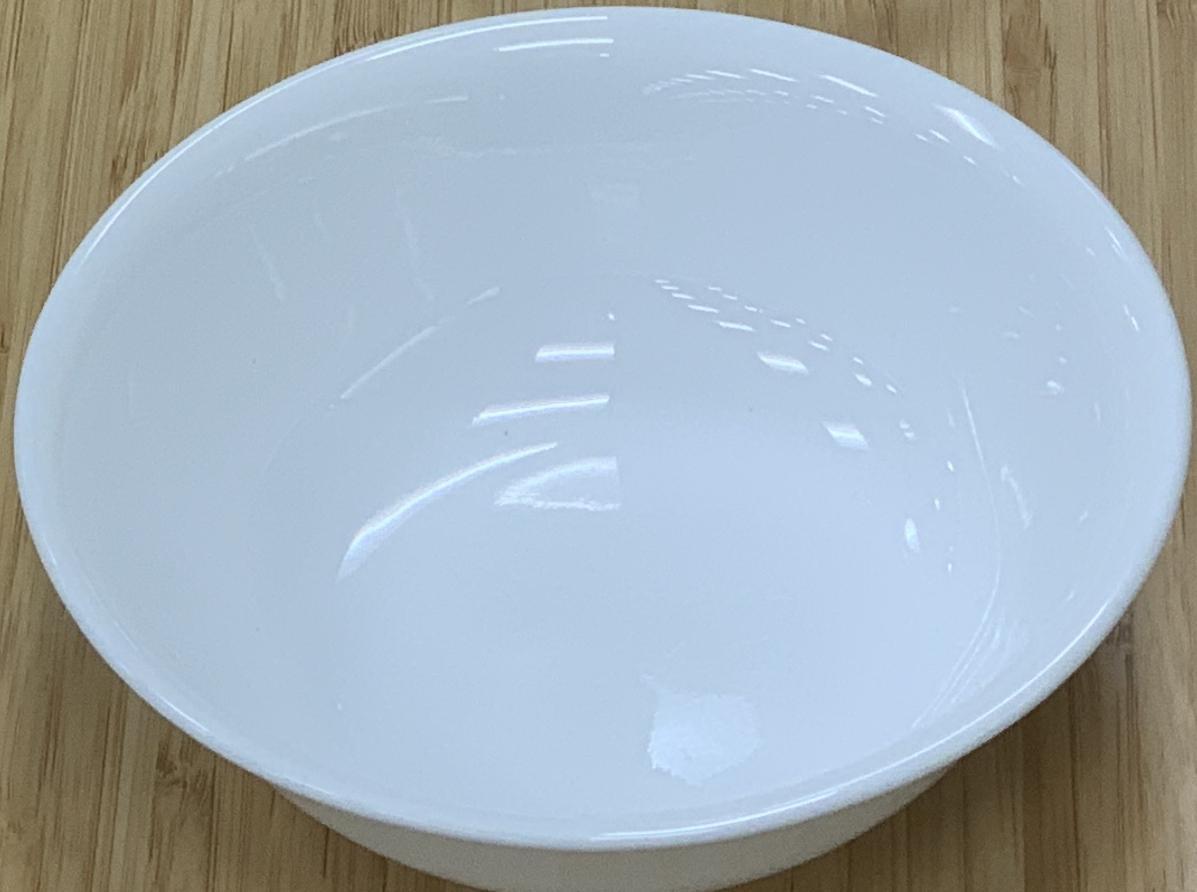}
        \caption{Ceramic bowl}%
        \label{fig:ceramic_bowl}
    \end{subfigure}
    \begin{subfigure}[t]{0.325\columnwidth}
        \includegraphics[height=3.65cm]{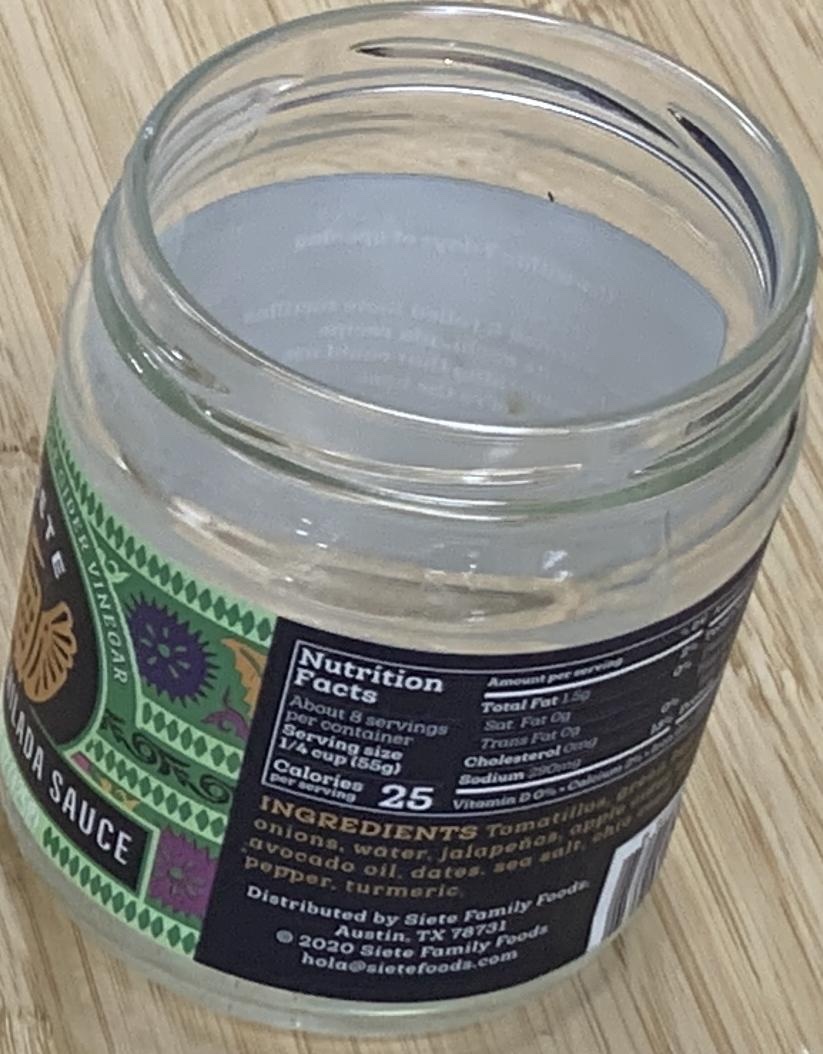}
        \caption{Glass jar}%
        \label{fig:glass_jar}
    \end{subfigure}
    \vspace{-7pt}
    \caption{\centering Objects from our real robot evaluation \vspace{-3pt}}%
\end{figure}

We find that using PG-InstructBLIP leads to successful robot executions more often than InstructBLIP. For example, when asked ``Is this object not plastic?" about the ceramic bowl in \cref{fig:ceramic_bowl}, InstructBLIP incorrectly assigns a likelihood of 0.89 to \emph{yes}, while PG-InstructBLIP only assigns 0.18. However, when asked ``Is this object translucent?" about the glass jar in \cref{fig:glass_jar}, both InstructBLIP and PG-InstructBLIP incorrectly assign likelihoods of 0.95 and 0.91 to \emph{yes}, respectively. We note that while these questions relate to physical concepts in \dataset, neither are formatted like the training questions for PG-InstructBLIP. For example, the training prompt for \emph{transparency} was ``Is this object transparent, translucent, or opaque?". This suggests that despite using a large pre-trained VLM, PG-InstructBLIP may sometimes still fail due to out-of-distribution questions.
We provide more results and visualizations on our \href{\website}{website}.

\begin{table}[t]
    \vspace{5pt}
    \begin{tabular}{p{3cm}C{1.5cm}C{2.6cm}}
        \toprule
                     & InstructBLIP & PG-InstructBLIP (ours) \\ 
        \midrule
        Scene 1      & 2/5          & \textbf{5/5}           \\
        Scene 2      & 2/5          & \textbf{4/5}           \\
        \midrule
        Overall      & 4/10         & \textbf{9/10}          \\
        \bottomrule
    \end{tabular}
    \vspace{-3pt}
    \caption{Success rates for real robot evaluation}%
    \label{table:real_results}
\end{table}

\section{Discussion}
\label{sec:discussion}

\noindent{\textbf{Summary.}}
In this work, we propose \dataset, the first large-scale dataset of physical concept annotations of real household object images, and demonstrate that fine-tuning a VLM on it significantly improves its physical reasoning abilities, including on held-out physical concepts. We find that using the fine-tuned VLM for real-world robotic planning improves performance on tasks that require physical reasoning. We believe our work makes progress toward expanding the applicability of VLMs for robotics.

\noindent{\textbf{Limitations and Future Work.}}
While we show \dataset can improve the physical reasoning of a VLM, it still makes errors relative to human judgment. Also, while our proposed methodology for continuous concepts improves relational grounding, which we show can be useful for robotic planning, the model outputs are not grounded in real physical quantities, which would be needed for some applications, e.g., identifying if an object is too heavy to be picked up. Future work can investigate incorporating data with real physical measurements to improve grounding.

While we believe the physical concepts in this work to have broad relevance for robotics, future work can expand on these for greater downstream applications. This could include expanding beyond physical reasoning, such as geometric reasoning (e.g., whether an object can fit inside a container), or social reasoning (e.g., what is acceptable to move off a table for cleaning). We believe our dataset is a first step towards this direction of using VLMs for more sophisticated reasoning in robotics.

\vspace{-5pt}
\section*{Acknowledgments}
This work was supported by NSF Awards 2132847, 1941722, and 2338203, ONR N00014-23-1-2355 and YIP, DARPA YFA, and Ford. We thank Minae Kwon, Siddharth Karamcheti, Suvir Mirchandani, and other ILIAD lab members for helpful discussions and feedback, and Siddharth Karamcheti for helping to set up the real robot evaluation.

\bibliographystyle{unsrt}
\bibliography{bibliography}

\clearpage

\appendix
\subsection{Physical Concepts Details}
\label{sec:phys_details}

In this section, we provide details on how we define each of our ten physical concepts, which we communicate to crowd-workers before annotation. We also list the pre-defined options for categorical concepts. \\

\noindent \textit{\textbf{Continuous-Valued, Applicable to All Objects}}

\noindent \textbf{Mass:} This refers to how heavy an object is. If an object has contents inside, this includes how heavy both the object and its contents are combined.

\noindent \textbf{Fragility:} This refers to how easily an object can be broken or damaged. An object has higher \emph{fragility} than another if a person would handle it more carefully to avoid breaking it.

\noindent \textbf{Deformability:} This refers to how easily an object can change shape without breaking. An object has more \emph{deformability} than another if less force is needed to change its shape without breaking it.

\noindent \textbf{Density (\emph{held-out}):} This refers to the amount of mass per unit of volume of the object. If an object has contents inside, this only refers to the object, not the contents. \\

\noindent \textit{\textbf{Continuous-Valued, Applicable to Containers}}

\noindent \textbf{Liquid Capacity (\emph{held-out}):} This refers to the volume of liquid a container can contain without spilling. \\

\noindent \textit{\textbf{Categorical-Valued, Applicable to All Objects}}

\noindent \textbf{Material:} This refers to what an object is made of. If an object is made of multiple materials, it refers to what material makes up the largest portion of the object that is visible. This does not refer to the contents of a container. The pre-defined options we include are \emph{plastic}, \emph{glass}, \emph{ceramic}, \emph{metal}, \emph{wood}, \emph{paper}, \emph{fabric}, \emph{food}, \emph{unknown}, and \emph{other} (annotator provides an open-ended response if this option is chosen).

\noindent \textbf{Transparency:} This refers to how much can be seen through an object. The pre-defined options we include are \emph{transparent}, \emph{translucent}, \emph{opaque}, and \emph{unknown}. \emph{Transparent} refers to an object that can be seen clearly through, almost as if it was not there. \emph{Translucent} refers to an object where some details can be seen through the object, but the details are not as clear as if it was \emph{transparent}. \emph{Opaque} refers to an object that cannot be seen through at all. This concept only refers to the object itself, and not the contents of a container. If different parts of an object have different levels of \emph{transparency}, it refers what level applies to the largest visible portion of the object. \\

\noindent \textit{\textbf{Categorical-Valued, Applicable to Containers}}

\noindent \textbf{Contents:} This refers to the contents of a container that are clearly visible and identifiable. The pre-defined options we include are \emph{nothing}, \emph{water}, \emph{food}, \emph{oil}, \emph{soap}, \emph{unknown}, and \emph{other} (annotator provides an open-ended response if this option is chosen).

\noindent \textbf{Can Contain Liquid:} This refers to if a container can be used to transport a liquid across a room without a person needing to be particularly careful about not spilling it. The pre-defined options we include are \emph{yes}, \emph{no}, and \emph{unknown}.

\noindent \textbf{Is Sealed:} This refers to if a container can be rotated by any amount in any direction without spilling its contents. The pre-defined options we include are \emph{yes}, \emph{no}, and \emph{unknown}. \\

\noindent \textbf{Container Categories.}
We define the following object categories from EgoObjects as containers: \emph{bottle}, \emph{container}, \emph{plate}, \emph{bowl}, \emph{mug}, \emph{water glass}, \emph{measuring cup}, \emph{wine glass}, \emph{tea cup}, \emph{frying pan}, \emph{flowerpot}, \emph{tin can}, \emph{kettle}, \emph{vase}, \emph{coffee cup}, \emph{mixing bowl}, \emph{saucer}, \emph{jug}, \emph{serving tray}, \emph{pitcher (container)}, and \emph{picnic basket}.

\vspace{-4pt}
\subsection{Automatic Annotation Details}
\label{sec:auto_details}

We list the object categories we assign to \emph{high} and \emph{low} tiers for automating preference pair annotations for continuous concepts in \cref{table:auto_continuous}. We list the object categories for which we assign a concept label in \cref{table:auto_categorical}. If a concept is not listed in these tables, we do not provide automatic annotations for that concept.

\begin{table*}
    \centering
    \begin{tabular}{l|p{5.25cm}|p{5.25cm}}
        \toprule
        \textbf{Concept} & \textbf{High} & \textbf{Low} \\
        \midrule
        Mass             & television, microwave oven, table, nightstand, chest of drawers & pen, paper, spoon, fork, glasses, sunglasses, scissors, watch, necklace, house/car key, pencil, earrings, ring, screwdriver, book, container, plate, bowl, pillow, remote control, clothing, mug, laptop, knife, mobile phone, toy, computer mouse, water glass, towel, headphones, spatula, frying pan, measuring cup, banana, wallet, blanket, candle, apple, wine glass, picture frame, computer keyboard, game controller/pad, tea cup, tin can, handbag, whisk, orange, belt, plastic bag, salt and pepper shakers, cutting board, perfume, stapler, footwear, tablet coputer, teddy bear, cookie, scarf, coffee cup, ball, mixing bowl, pear, alarm clock, light switch, bread, jacket, tennis ball, sandal, saucer, laptop charger, camera, yoga mat, power plugs and sockets, cream, shirt, baseball bat, sun hat, paper towel, kitchen knife, doll, can opener, sock, facial tissue holder, boot, hair dryer \\ \hline
        Fragility        & water glass, television & house/car key, dumbbell, screwdriver, kitchen knife \\ \hline
        Deformability    & pillow, clothing, towel, blanket, belt, plastic bag, scarf, jacket, yoga mat, shirt, paper towel, sock & remote control, mug, mobile phone, computer mouse, water glass, frying pan, flowerpot, scissors, wine glass, house/car key, dumbbell, cutting board, microwave oven, toaster, blender, pressure cooker, kitchen knife, table, spoon, laptop, knife, fork, glasses, spatula, sunglasses, chair, measuring cup, pencil, picture frame, computer keyboard, game controller/pad, tea cup, tin can, salt and pepper shakers, television, coffeemaker, stapler, tablet computer, kettle, vase, coffee cup, mixing bowl, computer monitor, stool, ring, alarm clock, light switch, saucer, printer, screwdriver, guitar, camera, jug, gas stove, baseball bat, humidifier, chest of drawers, sink, can opener, nightstand, hair dryer \\
        \bottomrule
    \end{tabular}
    \caption{Object category assignments to \emph{high} and \emph{low} tiers for continuous concepts}%
    \label{table:auto_continuous}
\end{table*}

\begin{table*}[ht]
    \centering
    \begin{tabular}{l|l|p{8cm}}
        \toprule
        \textbf{Concept}    & \textbf{Label} & \textbf{Categories} \\
        \midrule
        Material            & Plastic        & remote control, computer mouse, computer keyboard, game controller/pad, plastic bag \\ \cline{2-3} 
                            & Glass          & water glass, wine glass \\ \cline{2-3} 
                            & Metal          & tin can, kitchen knife, can opener \\ \cline{2-3} 
                            & Paper          & book, paper, paper towel \\ \cline{2-3} 
                            & Fabric         & clothing, towel, blanket, scarf, sock \\ \cline{2-3} 
                            & Food           & banana, apple, orange, cookie, pear, bread \\ \hline
        Transparency        & Transparent    & wine glass \\ \cline{2-3} 
                            & Opaque         & book, pillow, remote control, clothing, laptop, mobile phone, towel, headphones, spatula, chair, frying pan, banana, wallet, flowerpot, scissors, apple, houseplant, house/car key, pencil, computer keyboard, tin can, whisk, dumbbell, orange, belt, cutting board, toaster, teddy bear, tablet computer, cookie, pear, computer monitor, stool, light switch, bread, pressure cooker, scarf, laptop charger, guitar, camera, yoga mat, shirt, baseball bat, paper towel, kitchen knife, sink, chest of drawers, can opener, boot, nightstand, hair dryer \\ \hline
        Can Contain Liquid  & Yes            & bottle, mug, water glass, measuring cup, wine glass, tea cup, kettle, coffee cup, mixing bowl, jug, pitcher (container), tin can \\ \cline{2-3} 
                            & No             & picnic basket, serving tray \\ \hline
        Is Sealed           & No             & plate, bowl, mug, water glass, measuring cup, wine glass, tea cup, frying pan, flowerpot, kettle, vase, coffee cup, mixing bowl, saucer, jug, serving tray, pitcher (container), picnic basket \\ 
        \bottomrule
    \end{tabular}
    \caption{Concept label assignments of object categories for categorical concepts \vspace{-15pt}}%
    \label{table:auto_categorical}
\end{table*}

We originally assigned the label \emph{metal} for \emph{material} to the object category \emph{house/car key}, but realized after crowd-sourcing that not all instances of this category should have been given this assignment. Therefore, we manually labeled these examples for \emph{material}, but still considered these to be automatic annotations for dataset purposes.

\vspace{-4pt}
\subsection{Crowd-Sourcing Details}
\label{sec:crowd_details}

\begin{figure}[ht]
    \includegraphics[width=\linewidth]{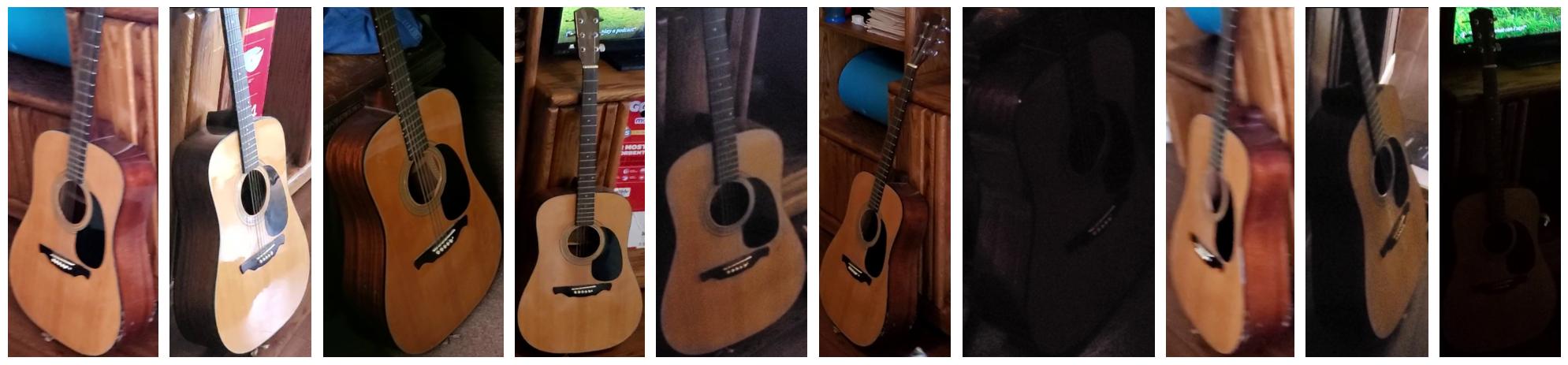}
    \centering
    \vspace{-15pt}
    \caption{\centering Bounding boxes for an instance of \emph{guitar}, in decreasing order of CLIP similarity \vspace{-10pt}}%
    \label{fig:bboxes}
\end{figure}

\noindent \textbf{Choosing Annotation Images.} There are multiple bounding box images in EgoObjects for each object instance. To determine which to present for annotating an object, we choose the bounding box with the highest CLIP \cite{Radford2021LearningTV} similarity with the object's category label, as a heuristic for the object's visibility. We use the CLIP-ViT-H-14-laion2B-s32B-b79K model from OpenCLIP \cite{ilharco_gabriel_2021_5143773}. In \cref{fig:bboxes}, we show an example of randomly sampled bounding boxes for an instance of the object category \emph{guitar}, arranged from left-to-right in decreasing order of CLIP similarity. The objects in bounding boxes with lower CLIP similarities tend to be less visible.

\noindent \textbf{Attention Checks.} We generate attention checks for crowd-workers by randomly sampling from the automatic annotations, which have known labels. For the concepts \emph{contents}, \emph{density}, and \emph{liquid capacity}, for which there are no automatic annotations, we manually label a small set of objects for attention checks.

\noindent \textbf{Other Details.}  Each annotation job on Prolific consisted of 250 annotations for a single concept, of which 25 are attention checks. Participants were paid an average of 15.50 US dollars per hour, and each annotation job took on average 20-30 minutes to complete, depending on the concept.

In the annotation user interface, for each object example, the object is shown in the context of its surrounding scene, with the object indicated by its bounding box. We also provide the object's category label to help clarify which object is to be annotated. Crowd-workers can choose an annotation label by clicking on an associated button, or typing an associated keyboard key. We also provide a \emph{back} option to go to the previous example to correct mistakes. For the concepts \emph{material} and \emph{contents}, the user may choose \emph{other} as an option, whereupon they are presented with a text box to type an open-ended label. We do not annotate objects from the categories \emph{pet}, \emph{cat}, and \emph{dog}, to omit objects that are living.

\begin{figure*}[t]
    \includegraphics[width=\linewidth]{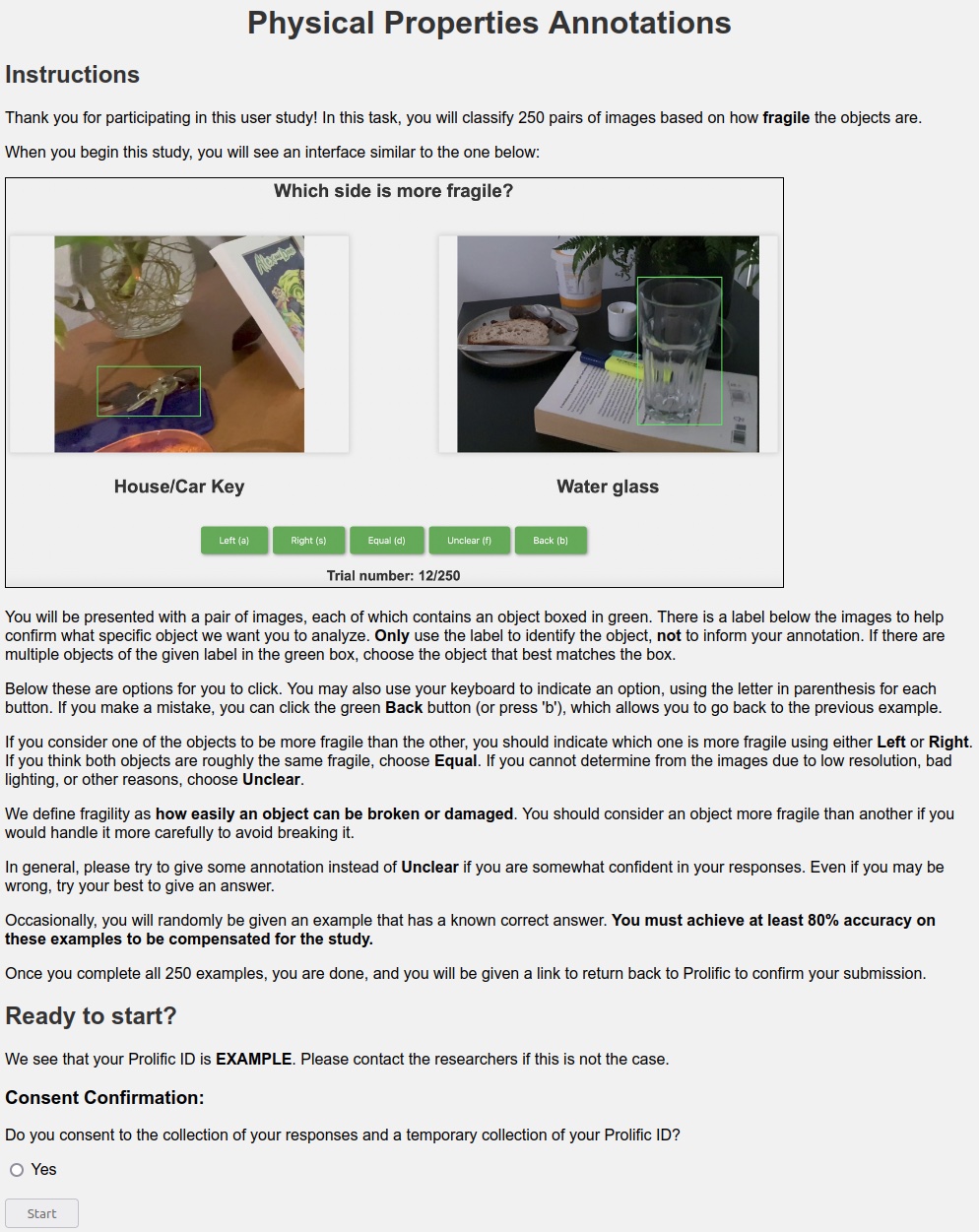}
    \centering
    \caption{Instruction page for the \emph{fragility} concept}%
    \label{fig:instruction}
\end{figure*}

We provide instructions to annotators that are specific to each concept, to encourage annotations that agree with our concept definitions. We provide an image of the instruction page provided to annotators for the \emph{fragility} concept, which also includes an example of the annotation user interface, in \cref{fig:instruction}. The instructions for how to annotate each property are also repeated at the bottom of the annotation user interface.

\section{Additional \dataset Statistics}
\label{sec:data_stat}
We detail the number of examples per concept and dataset split for \dataset in \cref{table:dataset_examples}. This is before any preprocessing of the data for annotator agreement or labels. For the crowd-sourced data, the count refers to the number of examples, not the number of annotations, for which there are three times as many. We also provide the percent of crowd-sourced examples with majority agreement (at least 2/3) and unanimous agreement per concept in \cref{table:dataset_agreement}.

\begin{table*}[ht]
    \centering
    \begin{tabular}{llrrr}
        \toprule
        \textbf{Concept}                  & \textbf{Source} & \textbf{Train}    & \textbf{Validation} & \textbf{Test} \\
        \midrule
        Mass                              & Crowd-sourced   & 2108              & 86                  & 56            \\
                                          & Automatic       & 87269             & 4536                & 2688          \\
        Fragility                         & Crowd-sourced   & 2096              & 99                  & 57            \\
                                          & Automatic       & 2397              & 110                 & 80            \\
        Deformability                     & Crowd-sourced   & 2101              & 84                  & 65            \\
                                          & Automatic       & 293540            & 13384               & 9888          \\
        Material                          & Crowd-sourced   & 2316              & 460                 & 374           \\
                                          & Automatic       & 612               & 130                 & 119           \\
        Transparency                      & Crowd-sourced   & 1993              & 394                 & 313           \\
                                          & Automatic       & 1046              & 224                 & 194           \\
        Contents                          & Crowd-sourced   & 641               & 134                 & 125           \\
                                          & Automatic       & 0                 & 0                   & 0             \\
        Can Contain Liquid                & Crowd-sourced   & 318               & 68                  & 64            \\
                                          & Automatic       & 342               & 70                  & 67            \\
        Is Sealed                         & Crowd-sourced   & 164               & 30                  & 31            \\
                                          & Automatic       & 444               & 91                  & 86            \\
        \midrule
        Density (\emph{held-out})         & Crowd-sourced   & 0                 & 0                   & 500           \\
        Liquid Capacity (\emph{held-out}) & Crowd-sourced   & 0                 & 0                   & 500           \\
        \bottomrule
    \end{tabular}
    \caption{Number of examples per concept and dataset split}%
    \label{table:dataset_examples}
\end{table*}

\begin{table*}[ht]
    \centering
    \begin{tabular}{lcc}
        \toprule
        \textbf{Concept}                  & \textbf{\% Majority Agreement} & \textbf{\% Unanimous Agreement} \\ 
        \midrule
        Mass                              & 94.2                           & 58.8                            \\
        Fragility                         & 93.6                           & 53.1                            \\
        Deformability                     & 90.5                           & 48.1                            \\
        Material                          & 93.7                           & 59.4                            \\
        Transparency                      & 97.0                           & 72.5                            \\
        Contents                          & 90.4                           & 49.8                            \\ 
        Can Contain Liquid                & 99.3                           & 64.2                            \\ 
        Is Sealed                         & 98.2                           & 74.7                            \\
        \midrule
        Density (\emph{held-out})         & 93.3                           & 50.7                            \\
        Liquid Capacity (\emph{held-out}) & 89.1                           & 46.0                            \\
        \bottomrule
    \end{tabular}
    \caption{Agreement among crowd-workers per concept \vspace{-15pt}}%
    \label{table:dataset_agreement}
\end{table*}

\vspace{-3pt}
\subsection{Training Details}
\label{sec:train_details}

\noindent \textbf{Hyperparameters.} We provide hyperparameters used for fine-tuning InstructBLIP in \cref{table:hparam}. These hyperparameters are largely derived from those proposed for fine-tuning BLIP-2 \cite{li2023blip}. When fine-tuning, we only update the Q-Former parameters, as done during instruction tuning for InstructBLIP. We use a linear warmup of the learning rate, followed by a linear decay with a minimum learning rate of 0. We fine-tune using mixed precision bfloat16 training. We use a prompt template for questions, which is used both during training and inference. We load the InstructBLIP model using the LAVIS library \cite{li2022lavis}. We train and evaluate using the evaluation image processor provided by LAVIS, as we do not use image data augmentation. 

\noindent \textbf{Validation \& Data Filtering.} For most experiments, we evaluate on validation data every 250 gradient steps and choose the checkpoint with the lowest validation loss. For experiments fine-tuning for a single concept, we validate every 100 gradient steps. Our validation set consists of all validation data for all concepts without balancing, except we limit the number of automatically generated examples for \emph{mass} and \emph{deformability} to 100. For validation data, we only use the bounding box image with the highest CLIP object category similarity score for each object, which for crowd-sourced data is also the bounding box image presented for annotation. For crowd-sourced validation data, we filter our data to only include examples with at least 2/3 majority agreement among annotators, and only use the majority label. We do not apply this filtering for training data. For preference pair annotations, we remove data annotated with \emph{unclear}.

\begin{table*}[ht]
    \centering
    \begin{tabular}{l|c}
        \toprule
        \textbf{Hyperparameter}     & \textbf{Value}                                                    \\
        \midrule
        Max fine-tuning steps       & 10000                                                             \\
        Warmup steps                & 1000                                                              \\
        Learning rate               & 1e-5                                                              \\
        Batch size                  & 128                                                               \\
        AdamW $\beta$               & (0.9, 0.999)                                                      \\
        Weight decay                & 0.05                                                              \\
        Image resolution            & 224                                                               \\
        Prompt template             & Question: \{\} Respond unknown if you are not sure. Short answer: \\
        \bottomrule
    \end{tabular}
    \caption{Hyperparameters for fine-tuning InstructBLIP}%
    \label{table:hparam}
\end{table*}

\begin{table*}[ht]
    \centering
    \begin{tabular}{l|l}
        \toprule
        \textbf{Concept}                  & \textbf{Question Prompt}                            \\
        \midrule
        Mass                              & Is this object heavy?                               \\ 
        Fragility                         & Is this object fragile?                             \\ 
        Deformability                     & Is this object deformable?                          \\
        Material                          & What material is this object made of?               \\
        Transparency                      & Is this object transparent, translucent, or opaque? \\
        Contents                          & What is inside this container?                      \\
        Can Contain Liquid                & Can this container hold a liquid inside easily?     \\
        Is Sealed                         & Is this container sealed?                           \\
        \midrule
        Density (\emph{held-out})         & Is this object dense?                               \\
        Liquid Capacity (\emph{held-out}) & Can this object hold a lot of liquid?               \\   
        \bottomrule
    \end{tabular}
    \caption{Question prompts for each concept, without object category labels \vspace{-15pt}}%
    \label{table:question_prompts}
\end{table*}

\noindent \textbf{Dataset Balancing.} We construct sub-datasets for dataset balancing purposes. For the categorical concepts except \emph{is sealed}, we combine the crowd-sourced and automatically annotated data for each concept into one sub-dataset per concept. For the other concepts, we keep separate sub-datasets for crowd-sourced and automatically annotated data. We keep separate sub-datasets for \emph{is sealed} because for its crowd-sourced data, we only train using the bounding box image for the object that was presented for annotation, rather than randomly sampling one of its bounding box images (as described in the below sub-section), as values for this concept may change for the same object instance. We keep separate datasets for the continuous concepts because there is a large imbalance between the number of crowd-sourced and automatically annotated examples for these concepts. To balance these sub-datasets, we sample from each of them during training at a rate proportional to the square root of the number of annotations in the sub-dataset, as proposed in InstructBLIP for instruction tuning.

\noindent \textbf{Additional Training Details.} For most objects, each time we sample one for training, we randomly sample one of its bounding box images as input to the model, as a form of data augmentation. We do not do this with crowd-sourced data for the \emph{contents} and \emph{is sealed} concepts, because labels for these concepts may vary across different images of the same object. Instead, we only use the bounding box image that was presented for annotation.

To promote robustness to different queries to the VLM, we include object category labels in the question prompt for half of the training examples (e.g., asking ``Is this \emph{bottle} heavy?"), and omit this information in the other half (e.g., asking ``Is this \emph{object} heavy?"). We experimented with training on one or multiple question prompts per concept, and found this to not significantly affect performance, so we only use one prompt per concept for simplicity. We include the question prompts for each concept in \cref{table:question_prompts}. These are versions of the prompts without object category labels. When including category labels, we replace either the word ``object" or ``container" with the object's category label from EgoObjects. We also pluralize the prompt to have correct grammar if the category label is plural.

We experimented with removing Q-Former text conditioning in InstructBLIP while fine-tuning, and found this to improve results on general VQA evaluation and evaluation with held-out paraphrased question prompts, so we report results using models trained without this text conditioning. In our ablation results in \cref{table:ablation_results}, we find that this does not significantly change performance for our main crowd-sourced evaluation.

\vspace{-4pt}
\subsection{Evaluation Details}
\label{sec:data_eval}

\noindent \textbf{Further \dataset Evaluation Details.} 
For crowd-sourced test evaluation data, we only include examples with at least 2/3 annotator agreement, and use the majority label as ground-truth. For categorical concepts, we predict by choosing the label with the highest likelihood out of all labels in \dataset for the concept. For continuous concepts, we predict the object in a pair with the higher score from \cref{pref_score} as the one with higher concept value. We only evaluate on preference examples with a definite, non-equal preference label. For the \emph{Most Common} baseline with continuous concepts, we also only include examples with a definite, non-equal preference when determining the most common label in the training data. We note that that \emph{Most Common} baseline is not particularly meaningful for continuous concepts, because the preference labels and predictions are invariant to ordering in each preference pair. Therefore, a more natural baseline for these concepts would be random guessing, which would achieve 50\% accuracy.

Similarly as with validation data, for test data we only evaluate using the bounding box image with the highest CLIP object category similarity per object, which for crowd-sourced data is also the bounding box image presented for annotation. We evaluate using the same question prompts per concept as during training, which are listed in \cref{table:question_prompts}. Unless stated otherwise, we report evaluation results without object category labels in the question prompt, because this gives slightly better results for the base InstructBLIP model.

\begin{table*}[ht]
    \centering
    \begin{tabular}{l|l}
        \toprule
        \textbf{Concept}   & \textbf{Question Prompt}                                               \\
        \midrule
        Mass               & Does this object weigh a lot?                                          \\ 
        Fragility          & Is this object easily breakable?                                       \\ 
        Deformability      & Is this object easily bendable?                                        \\
        Material           & What is this object made of?                                           \\
        Transparency       & Would you describe this object as opaque, transparent, or translucent? \\
        Contents           & What does this container contain?                                      \\
        Can Contain Liquid & Is this container able to hold water inside easily?                    \\
        Is Sealed          & Is this container sealed shut?                                         \\
        \bottomrule
    \end{tabular}
    \caption{Paraphrased question prompts for main concepts, without object category labels \vspace{-15pt}}%
    \label{table:paraphrase_prompts}
\end{table*}

\noindent \textbf{Text Only Baseline.}
For this baseline, we use ground truth object category labels from EgoObjects. We use the `text-davinci-003' InstructGPT model \cite{ouyang2022training} as our LLM. For each concept, we use 128 in-context examples randomly sampled from the training data in \dataset for that concept. Because in-context learning is limited by context length, and therefore it is desirable to use the best quality in-context examples when possible, we first apply to the training data the same majority filtering process used on crowd-sourced test data as described in the previous subsection. We also remove preference annotations with the label \emph{unclear}, as done in our VLM fine-tuning setup.
We treat each example as a question answering task, using question prompts for each concept similar to those in \cref{table:question_prompts}, but modified to refer to general object classes, rather than specific instances.  We make predictions by selecting the most likely completion of the LLM conditioned on the in-context examples and test example. For categorical concepts, we first include in the LLM context all possible labels in \dataset for the concept. For continuous concepts, because we only evaluate on examples with definite preferences, we restrict predictions to only definite preferences using logit bias, although the in-context examples may include \emph{equal} as a possible answer.

\noindent \textbf{Paraphrased Question Prompts.}
In \cref{table:paraphrase_prompts}, we list the paraphrased prompts used in the evaluation for \cref{table:paraphrase}.

\begin{table}[ht]
    \centering
    \begin{tabular}{lcc}
        \toprule
               & InstructBLIP & PG-InstructBLIP (ours) \\
        \midrule
        VQAv2  & 71.4         & 67.5                   \\
        OK-VQA & 52.4         & 48.7                   \\
        \bottomrule
    \end{tabular}
    \caption{Accuracy on existing VQA benchmarks \vspace{-5pt}}%
    \label{table:vqa_results}
\end{table}

\noindent \textbf{Limited VQA Degradation.}
Ideally, training on \dataset should be done while co-training on other vision and language datasets to preserve general reasoning abilities. In this work, we do not do this because we focus primarily on physical reasoning. However, we show that fine-tuning on only \dataset does not significantly degrade general VQA performance. In \cref{table:vqa_results}, we compare InstructBLIP to PG-InstructBLIP on VQAv2 \cite{goyal2017making} and OK-VQA \cite{marino2019ok}. These results suggest that existing systems using VLMs can benefit from \dataset for physical reasoning, without sacrificing other reasoning abilities.

We perform VQA evaluation using the LAVIS library, using their configurations for evaluation of BLIP-2. Although PG-InstructBLIP is fine-tuned without Q-Former text conditioning, we found that Q-Former text conditioning during VQA evaluation improved performance, so we report these results. We believe this is because InstructBLIP was instruction tuned with this text conditioning. We also experimented with VQA evaluation on PG-InstructBLIP fine-tuned with Q-Former text conditioning, but found this to have worse results, possibly due to overfitting on our limited variety of question prompts. We believe these issues can be mitigated by co-training on \dataset in combination with other vision and language datasets, which we leave for future work.

Motivated by these VQA results, for our planning evaluations we also evaluate PG-InstructBLIP using Q-Former text conditioning, to avoid possible degradation when answering questions that do not pertain concepts in \dataset. We verified that evaluating PG-InstructBLIP using Q-Former text conditioning did not significantly affect test accuracy on \dataset.

\noindent \textbf{Including Object Category Labels in Question Prompts.}
We generally report evaluation results without ground-truth object category labels in the question prompt. In \cref{table:cat_labels}, we compare including object category labels or not, and find that all models are not extremely sensitive to this.

\begin{table*}[ht]
    \centering
    \begin{tabular}{lC{1.5cm}C{1.5cm}C{1.5cm}C{1.5cm}C{1.5cm}C{1.5cm}}
        \toprule
                            & \multicolumn{2}{c}{InstructBLIP} & \multicolumn{2}{c}{Single Concept FT (ours)} &\multicolumn{2}{c}{PG-InstructBLIP (ours)} \\
        \midrule
        Category Labels     & Yes            & No              & Yes                  & No                    & Yes                & No                   \\
        \midrule
        Mass                & 60.0           & 62.2            & \textbf{84.4}        & 80.0                  & 80.0               & 80.0                 \\
        Fragility           & 75.7           & 78.4            & 91.2                 & 91.2                  & \textbf{97.3}      & 94.6                 \\
        Deformability       & 69.8           & 67.4            & 88.4                 & \textbf{95.3}         & 90.7               & 93.0                 \\
        Material            & 73.3           & 67.1            & \textbf{86.8}        & 83.7                  & 85.7               & 84.6                 \\
        Transparency        & 84.5           & 85.8            & 89.1                 & 89.4                  & 89.8               & \textbf{90.1}        \\
        Contents            & 34.2           & 35.1            & 80.7                 & 81.6                  & 82.5               & \textbf{83.3}        \\
        Can Contain Liquid  & 57.8           & 59.4            & 84.4                 & 84.4                  & 82.8               & \textbf{87.5}        \\
        Is Sealed           & 71.0           & 74.2            & 80.6                 & 80.6                  & \textbf{87.1}      & \textbf{87.1}        \\
        \midrule
        Average             & 65.8           & 66.2            & 85.7                 & 85.8                  & 87.0               & \textbf{87.5}        \\       
        \bottomrule
    \end{tabular}
    \caption{\centering Test accuracy for main concepts on crowd-sourced \dataset, with and without object category labels}%
    \label{table:cat_labels}
\end{table*}

\noindent \textbf{Including Concept Definitions in Question Prompts.}
While we did not spend extensive effort designing the question prompts for each concept (shown in \cref{table:question_prompts}), we aimed for them to be concise while still eliciting the desired concept. As seen in \cref{table:def_results}, the base InstructBLIP model achieves above chance performance on all concepts, suggesting that these prompts do elicit the desired concept to some extent. However, these prompts do not contain our definitions for each concept provided to annotators, as described in \cref{sec:phys_details}. We analyze whether including concept definitions in the question prompt would improve base VLM performance in \cref{table:def_results}, which contains our original crowd-sourced test accuracy results, with additional evaluation of the base InstructBLIP model using modified prompts that contain concept definitions, which we provide in \cref{table:def_prompts}.
We find that while including concept definitions improves performance for some concepts (\emph{mass}, \emph{deformability}, \emph{contents}, \emph{can contain liquid}), this still does not match PG-InstructBLIP on these concepts, and overall performance in fact \emph{decreases} compared to the original prompts. We believe this could be because InstructBLIP does not have strong enough language understanding to properly incorporate the concept definitions when providing responses. For this reason, and for simplicity, we use prompts without concept definitions in the rest of our experiments.

\begin{table*}[ht]
    \small
    \centering
    \begin{tabular}{l|p{10cm}}
        \toprule
        \textbf{Concept}   & \textbf{Question Prompt} \\
        \midrule
        Mass               & The heaviness of an object refers to its mass. It includes the contents of the object if it has something inside it. Is this object heavy? \\ \hline
        Fragility          & Fragility refers to how easily an object can be broken or damaged. Is this object fragile? \\ \hline
        Deformability      & Deformability refers to how easily an object can change shape without breaking. Is this object deformable? \\ \hline
        Material           & The material of an object refers to what material makes up the largest portion of the object that is visible. It does not refer to the contents of a container. What material is this object made of? \\ \hline
        Transparency       & Transparency refers to how much can be seen through an object. A transparent object can be clearly seen through, almost as if it was not there. A translucent object can be seen through with some details, but not as clearly as if it was transparent. An opaque object cannot be seen through at all. The transparency of an object does not refer to the transparency of its contents if it has anything inside it. Is this object transparent, translucent, or opaque? If different portions of the object have different levels of transparency, respond with the level that applies to the largest visible portion of the object. \\ \hline
        Contents           & What is inside this container? Only respond with contents that are clearly visible and identifiable. \\ \hline
        Can Contain Liquid & A container can contain liquid if it can be used to transport a liquid across a room without a person needing to be particularly careful about not spilling it. Can this container contain liquid? \\ \hline
        Is Sealed          & A container is sealed if it can be rotated by any amount in any direction without spilling its contents if it has anything inside. Is this container sealed? \\
        \bottomrule
    \end{tabular}
    \caption{Question prompts with definitions for each main concept, without object category labels}%
    \label{table:def_prompts}
\end{table*}

\begin{table*}[ht]
    \small
    \centering
    \begin{tabular}{lccc}
        \toprule
                            & \multicolumn{2}{c}{InstructBLIP}  & PG-InstructBLIP (ours) \\
        \midrule
        Prompt Type         & Original & w/ Concept Definitions & Original               \\
        \midrule
        Mass                & 62.2     & 71.1                   & \textbf{80.0}          \\
        Fragility           & 78.4     & 67.6                   & \textbf{94.6}          \\
        Deformability       & 67.4     & 69.8                   & \textbf{93.0}          \\
        Material            & 67.1     & 66.0                   & \textbf{84.6}          \\
        Transparency        & 85.8     & 65.3                   & \textbf{90.1}          \\
        Contents            & 35.1     & 36.0                   & \textbf{83.3}          \\
        Can Contain Liquid  & 59.4     & 64.1                   & \textbf{87.5}          \\
        Is Sealed           & 74.2     & 64.5                   & \textbf{87.1}          \\
        \midrule
        Average             & 66.2     & 63.0                   & \textbf{87.5}          \\   
        \bottomrule
    \end{tabular}
    \caption{\centering Test accuracy for main concepts on crowd-sourced \dataset, with additional base InstructBLIP evaluation on prompts with definitions for each concept}%
    \label{table:def_results}
\end{table*}

\begin{table*}[ht]
    \centering
    \begin{tabular}{lcccc}
        \toprule
                            & \multicolumn{2}{c}{InstructBLIP} & \multicolumn{2}{c}{PG-InstructBLIP (ours)} \\
        \midrule
        LLM Version         & Flan-T5 XL     & Flan-T5 XXL     & Flan-T5 XL          & Flan-T5 XXL          \\
        \midrule
        Mass                & 62.2           & 62.2            & \textbf{82.2}       & 80.0                 \\
        Fragility           & 64.9           & 78.4            & \textbf{97.3}       & 94.6                 \\
        Deformability       & 48.8           & 67.4            & \textbf{97.7}       & 93.0                 \\
        Material            & 69.1           & 67.1            & 82.6                & \textbf{84.6}        \\
        Transparency        & 74.0           & 85.8            & 87.5                & \textbf{90.1}        \\
        Contents            & 18.4           & 35.1            & \textbf{86.8}       & 83.3                 \\
        Can Contain Liquid  & 68.8           & 59.4            & \textbf{87.5}       & \textbf{87.5}        \\
        Is Sealed           & 67.7           & 74.2            & 80.6                & \textbf{87.1}        \\
        \midrule
        Average             & 59.2           & 66.2            & \textbf{87.8}       & 87.5                 \\   
        \bottomrule
    \end{tabular}
    \caption{Test accuracy for main concepts on crowd-sourced \dataset, using different VLM versions \vspace{-15pt}}%
    \label{table:smaller_vlm}
\end{table*}

\noindent \textbf{Using a Smaller VLM.} To analyze the effect of VLM size on physical reasoning, 
in \cref{table:smaller_vlm} we provide evaluation results using the InstructBLIP version with the smaller Flan-T5 XL as its base LLM, compared to the Flan-T5 XXL version used in all other experiments. We find that while the smaller Flan-T5 XL version generally has worse base performance, after fine-tuning on \dataset, we see that performance is comparable between the two model sizes. This suggests that for physical reasoning, fine-tuning on human data such as \dataset could reduce the need for larger model sizes. While fine-tuned evaluation performance is similar across model sizes, for simplicity of comparison, we only report results using the larger Flan-T5 XXL models in all other experiments.

\begin{table}[ht]
    \centering
    \begin{adjustbox}{width=\textwidth}
    \begin{tabular}{lcccc}
        \toprule
                            & InstructBLIP & PG-InstructBLIP (ours) \\
        \midrule
        Mass                & 72.8         & \textbf{99.9}          \\
        Fragility           & 95.0         & \textbf{100}           \\
        Deformability       & 96.0         & \textbf{98.8}          \\
        Material            & 89.1         & \textbf{98.3}          \\
        Transparency        & 97.4         & \textbf{100}           \\
        Can Contain Liquid  & 98.5         & \textbf{100}           \\
        Is Sealed           & \textbf{100} & \textbf{100}           \\
        \midrule
        Average             & 92.7         & \textbf{99.6}          \\
        \bottomrule
    \end{tabular}
    \end{adjustbox}
    \caption{\centering Test accuracy for main concepts on automatically annotated \dataset \vspace{-5pt}}%
    \label{table:auto_results}
\end{table}

\noindent \textbf{Results on Automatically Annotated Data.} We report evaluation results on automatically annotated data in \cref{table:auto_results}. Performance is generally much higher on this data compared to the crowd-sourced data, because these are easier examples that can be determined from object categories alone.

\begin{figure}[ht]
    \centering
    \includegraphics[width=\linewidth]{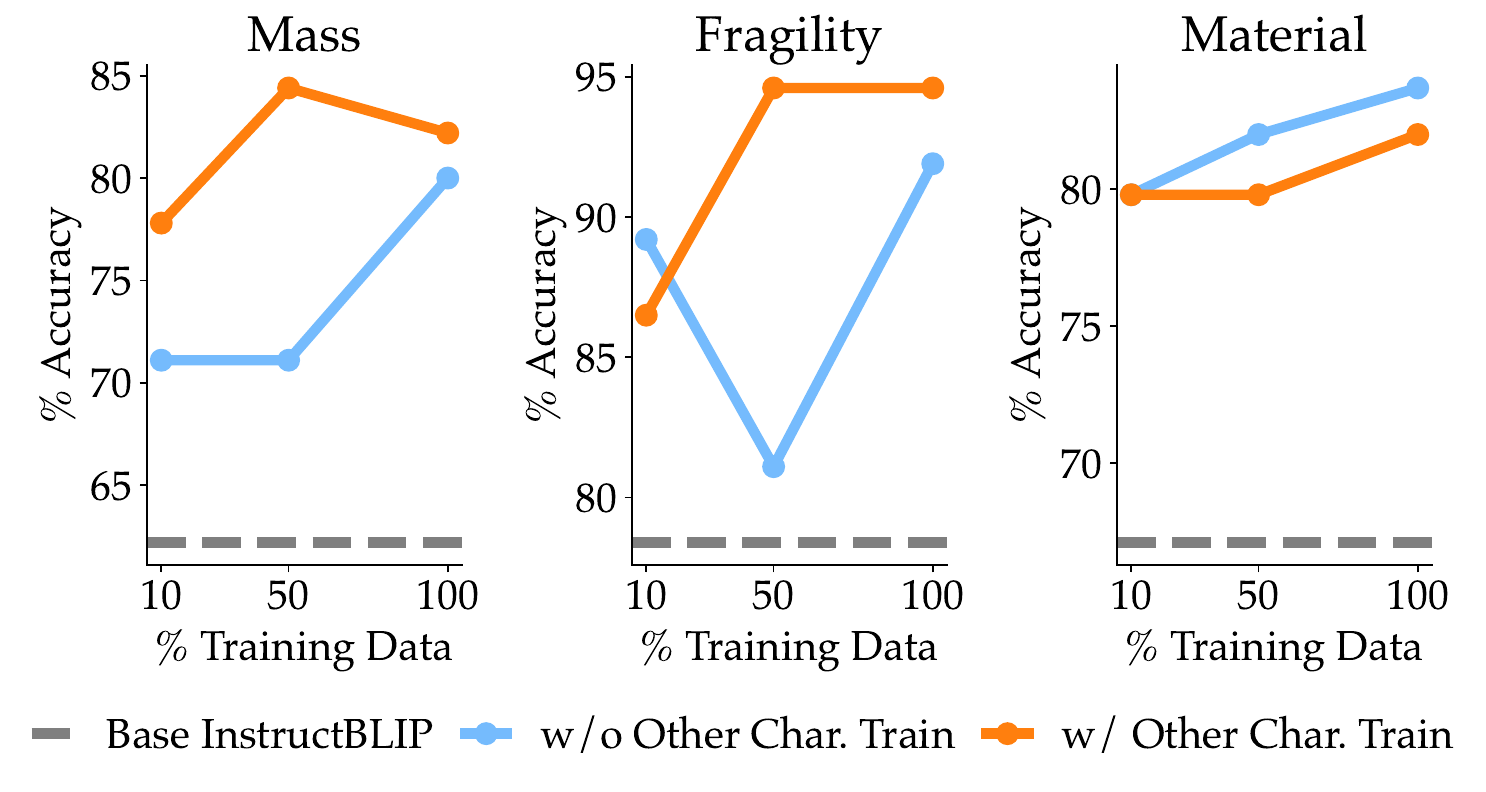}%
    \caption{Performance scaling on held-out concepts}%
    \label{fig:dataset_heldout}
\end{figure}

\noindent \textbf{Held-Out Concept Scaling.} In these experiments, we evaluate the transfer abilities of InstructBLIP across different concepts when fine-tuning on \dataset. We fine-tune models on data from \dataset for all concepts except one, and then report results of additional fine-tuning on the held-out concept. We compare to fine-tuning base InstructBLIP without training on the other concepts, and base InstructBLIP without any fine-tuning. Results for three concepts are shown in \cref{fig:dataset_heldout}. We chose these concepts because we believed they had the most generalization potential from the other concepts. We find that there are some signs of positive transfer on \emph{mass} and \emph{fragility}, although we see slight negative transfer on \emph{material}. We believe that more positive transfer could be attained by co-training with other vision and language datasets.

\begin{table*}[ht]
    \centering
    \begin{adjustbox}{width=\textwidth}
    \begin{tabular}{lccccccc}
        \toprule
                            & PG-InstructBLIP (ours) & No Auto Data  & Filtered & Q-Former Text & No Category Info & Only Category Info \\
        \midrule
        Mass                & 80.0                   & \textbf{84.4} & 75.6     & 80.0          & 75.6             & 77.8               \\
        Fragility           & 94.6                   & 94.6          & 97.3     & 97.3          & \textbf{100}     & \textbf{100}       \\
        Deformability       & \textbf{93.0}          & \textbf{93.0} & 90.7     & 90.7          & \textbf{93.0}    & 88.4 \\
        Material            & 84.6                   & 83.4          & 83.4     & 85.4          & 84.6             & \textbf{86.5}      \\
        Transparency        & 90.1                   & 89.4          & 89.1     & \textbf{92.1} & 89.8             & 91.7               \\
        Contents            & 83.3                   & 81.6          & 85.1     & \textbf{87.7} & 84.2             & 86.8               \\
        Can Contain Liquid  & 87.5                   & 89.1          & 85.9     & 84.4          & \textbf{90.6}    & 84.4               \\
        Is Sealed           & \textbf{87.1}          & 83.9          & 83.9     & 71.0          & 80.6             & \textbf{87.1}      \\
        \midrule
        Average             & 87.5                   & 87.4          & 86.4     & 86.1          & 87.3             & \textbf{87.8}      \\
        \bottomrule
    \end{tabular}
    \end{adjustbox}
    \caption{Ablation results for main concepts on crowd-sourced \dataset \vspace{-15pt}}%
    \label{table:ablation_results}
\end{table*}

\noindent \textbf{Ablations.}
We report additional ablation results on crowd-sourced \dataset examples in \cref{table:ablation_results}. We list each ablation below:
\begin{enumerate}
    \item \emph{No Auto Data}: Instead of training on both crowd-sourced and automatically annotated data, we train on only crowd-sourced data.
    \item \emph{Filtered}: Instead of training on all annotations for crowd-sourced data, we filter the data similarly as during evaluation: we only include examples with at least 2/3 annotator agreement, and use the majority label as ground-truth.
    \item \emph{Q-Former Text}: Instead of removing Q-Former text conditioning during fine-tuning, we include it, as done for the original InstructBLIP model.
    \item \emph{No Category Info}: Instead of training on both question prompts with and without object category information, we only train on question prompts \emph{without} it.
    \item \emph{Only Category Info}: Instead of training on both question prompts with and without object category information, we only train on question prompts \emph{with} it. Here, unlike the rest of the evaluations, we evaluate with object category information to match the training setup.
\end{enumerate}

We find that overall performance for each ablated version of our model does not change significantly, suggesting some robustness of our fine-tuning process to different design decisions. In particular, we find that including automatically annotated data does not significantly impact performance on crowd-sourced data, which perhaps is not surprising because base InstructBLIP already performs well on automatically annotated examples, as seen in \cref{table:auto_results}. \emph{Only Category Info} very slightly improves upon PG-InstructBLIP, but uses privileged object category information at evaluation time.

\vspace{-4pt}
\subsection{Real Scene Planning Evaluation Details}
\label{sec:plan_eval}

\noindent \textbf{Planning Framework.} Our planning framework consists of first providing the scene image to an OWL-ViT ViT-L/14 open-vocabulary object detector \cite{minderer2022simple}, which produces object bounding boxes and category labels from the EgoObjects categories. We then provide the list of detected objects and the task instruction to our LLM, which is GPT-4 \cite{openai2023gpt4} with temperature 0. The LLM is additionally provided with the robotic primitives, and a few-shot chain-of-thought prompt \cite{wei2022chain} with instructions to ask questions about objects in the scene to determine how to complete the task, and then produce a plan using the primitives. There is no constraint on the questions that the LLM can ask, except for encouragement in the prompt to ask questions that can be answered with \emph{yes} or \emph{no}. The same prompt is used for all scenes and tasks, which we provide in \cref{lst:vlm_prompt}. 

After the LLM asks a set of object-centric questions, a VLM answers each question prompted with the bounding box of the object indicated by the LLM, and then provides the LLM with its highest likelihood responses and their associated likelihoods/confidence scores, as done in prior work for VQA \cite{shao2023prompting}. This continues until the LLM decides it has enough information, whereupon it either indicates that the task is not possible, or produces a plan consisting of a list of primitives to execute for the task. The few-shot examples in \cref{lst:vlm_prompt} illustrate how interaction between the LLM and VLM for planning is structured.

\noindent \textbf{Primitives.}
We list the primitives for our real scene planning evaluation below:
\begin{itemize}
  \item go to object [X]
  \item pick up object [X]
  \item bring to human object [X]
  \item put down object [X]
  \item done
\end{itemize}
The primitives (except \emph{done}) are parameterized by a letter (in place of [X]) that identifies each detected object in the scene. The assignment of letters is provided in the list of object detections given to the LLM planner.

\noindent \textbf{Scenes and Tasks.}
In \cref{table:plan_scenes}, we provide the scene images in our evaluation, and the detected objects and task instructions for each scene. We also indicate the task type for each instruction.

\noindent \textbf{Prompts.}
We provide the prompts used by our LLM-based planning framework for our scene planning evaluation. The version with VLM interaction is in \cref{lst:vlm_prompt} and the version without VLM interaction is in \cref{lst:no_vlm_prompt}. The parts of the prompts in square brackets are replaced with the corresponding information specific to the task, in the same format as the prompt example.

\begin{lstlisting}[basicstyle=\ttfamily\scriptsize, backgroundcolor = \color{lightgray},
breaklines=true, caption={Prompt for LLM planner with VLM interaction.}, label={lst:vlm_prompt}]
You are a household robot. You are able to move most household objects, but not large or heavy furniture. You are to be safe and not break anything.
You will be given a list of objects in the scene.
A human will give you a task instruction to perform.
First, ask questions about the objects to learn more about them to determine how to properly complete the task.
Indicate the object letters before asking the question, and do not provide the answer.
Format each question like "Question about object [A, B]: Is this object heavy?".
Only ask informative questions that help you understand how to properly complete the task.
Some of the object descriptions in the list may be inaccurate, so it is good to ask questions to confirm information about them.
Ask questions that can be answered with yes or no when possible.
Only ask one question at a time.
After each question, you will be given a response containing the 3 most likely possible answers for each object, with corresponding confidence scores.
An example response for one object would be "A: yes (0.8), no (0.1), unknown (0.1)".
Otherwise, use the information you have obtained to produce a numbered list of actions to perform the task, starting with "Plan:".
You are only able to perform the following actions, where "X" is a placeholder for an object letter from the given list of objects:
1. go to object X
2. pick up object X
3. bring to human object X
4. put down object X
5. done

Only perform actions that are possible. You are able to move most household objects, including most electronic appliances, but not large or heavy furniture or parts of a room (like a door or heater). You are to be safe and not break anything. Choose the "done" action when the plan is complete and then stop. If no plan can satisfy the task, tell the human, choose the "done" action and stop.
Only respond with the plan list and do not include any other text.

Scene 1:
The following objects are in the scene: A (bottle), B (bowl), C (countertop), D (shirt), E (lock)

Instruction: Bring me the heaviest object.

Thought: I can ask which object is heavy out of all the objects, and I can find the one with the largest weight that I can carry.
Question about [A, B, C, D, E]: Is this object heavy?
Answer:
A: No (0.50), Yes (0.24), Unknown (0.21)
B: No (0.90), Unknown (0.05), Yes (0.04)
C: Yes (0.80), Unknown (0.10), No (0.05)
D: No (0.73), Unknown (0.20), Yes (0.06)
E: No (0.41), Yes (0.35), Unknown (0.19)
Thought: The heaviest object is the one with the highest Yes value or the lowest No value (if none are Yes). The heaviest object is C. However, I cannot carry a countertop since it is a heavy piece of furniture. Therefore, I will to choose the next heaviest, which would be E, a lock that I can carry easily.
Plan:
1. Go to object E
2. Pick up object E
3. Bring to human object E
4. Done

Instruction: Bring me the most deformable container.
Thought: I can first ask which objects are containers out of all the objects I can carry. Of these containers, I can ask which is deformable.
Question about [A, B, C, D, E]: Is this object a container?
Answer:
A: Yes (0.55), No (0.20), Unknown (0.19)
B: Yes (0.90), Unknown (0.05), No (0.04)
C: No (0.85), Unknown (0.06), Yes (0.05)
D: No (0.62), Unknown (0.20), Yes (0.06)
E: No (0.41), Yes (0.35), Unknown (0.19)
Thought: The only objects that are confidently Yes are B and A. I should ask which is deformable next.
Question about [A, B]: Is this object deformable?
Answer:
A: Yes (0.80), Unknown (0.15), No (0.04)
B: No (0.55), Unknown (0.26), Yes (0.05)
Thought: The most deformable object is the one with the highest Yes value or the lowest No value (if none are Yes). Since some of the answers are Yes, the answer is A (the highest yes value).
Plan:
1. Go to object A
2. Pick up object A
3. Bring to human object A
4. Done

Scene 2:
The following objects are in the scene: [list of objects in the scene]

Instruction: [instruction specified here]
\end{lstlisting}

\begin{lstlisting}[basicstyle=\ttfamily\scriptsize, backgroundcolor = \color{lightgray},
breaklines=true, caption={Prompt for LLM planner without VLM interaction.}, label={lst:no_vlm_prompt}]
You are a household robot. You are able to move most household objects, but not large or heavy furniture. You are to be safe and not break anything.
You will be given a list of objects in the scene.
A human will give you a task instruction to perform.
Use the object information to produce a numbered list of actions to perform the task, starting with "Plan:".
You are only able to perform the following actions, where "X" is a placeholder for an object letter from the given list of objects:
1. go to object X
2. pick up object X
3. bring to human object X
4. put down object X
5. done

Only perform actions that are possible. You are able to move most household objects, including most electronic appliances, but not large or heavy furniture or parts of a room (like a door or heater). You are to be safe and not break anything. Choose the "done" action when the plan is complete and then stop. If no plan can satify the task, tell the human, choose the "done" action and stop.
Only respond with the plan list and do not include any other text.

Scene 1:
The following objects are in the scene: A (bottle), B (bowl), C (countertop), D (shirt), E (lock)

Instruction: Bring me the heaviest object.

Thought: I cannot carry a countertop since it is a heavy piece of furniture. Out of the rest, a good guess would be the lock.
Plan:
1. Go to object E
2. Pick up object E
3. Bring to human object E
4. Done

Instruction: Bring me the most deformable container.
Thought: Typically shirts are easy to fold, so a good choice for the most deformable object would be the shirt.
Plan:
1. Go to object D
2. Pick up object D
3. Bring to human object D
4. Done

Scene 2:
The following objects are in the scene: [list of objects in the scene]

Instruction: [instruction specified here]
\end{lstlisting}

\noindent \textbf{Evaluation Procedure.}
We evaluate task planning accuracy using a non-author human evaluator. For each evaluation, the evaluator is given the task instruction, the image of the scene, the list of detected objects in the scene and their bounding boxes, and the generated task plan, and they are asked to evaluate whether the task plan successfully performed the task instruction for the given scene. We provide the following instructions to the evaluator on what to consider when evaluating whether a plan was correct:

\noindent\fbox{%
    \parbox{0.47\textwidth}{%
        Instructions: For each scene, there is a list of objects (under `Options:'). Below that is a table of tasks for that scene. The instruction given to a robot is on the left. On the right are the choices from 3 different robots. You need to mark which ones are correct or incorrect. It may be possible that multiple robots got it right or none of them got it right. Be aware that in tasks that involve moving objects, the robot should not plan to move an object that is very heavy, like large furniture.
    }%
} \\

While the planner usually creates plans using only the provided primitives, it sometimes specifies primitives that were not provided. Because the purpose of this evaluation is on assessing if a LLM planner can benefit from physical reasoning using a VLM, and not on creating a functional planning system, we do not do anything to handle these cases. We the evaluator to judge if these plans satisfy the task instruction like the others. We provide example executions for different versions of our planning framework on our \href{\website}{website}.

\begin{table*}[ht]
    \centering
    \begin{tabular}{p{0.3\textwidth}p{0.3\textwidth}p{0.3\textwidth}}
        \toprule
        \textbf{Scene Image} & \textbf{Object Detections} & \textbf{Task Instructions} \\
        \midrule
        \raisebox{-\totalheight}{\includegraphics[width=0.3\textwidth]{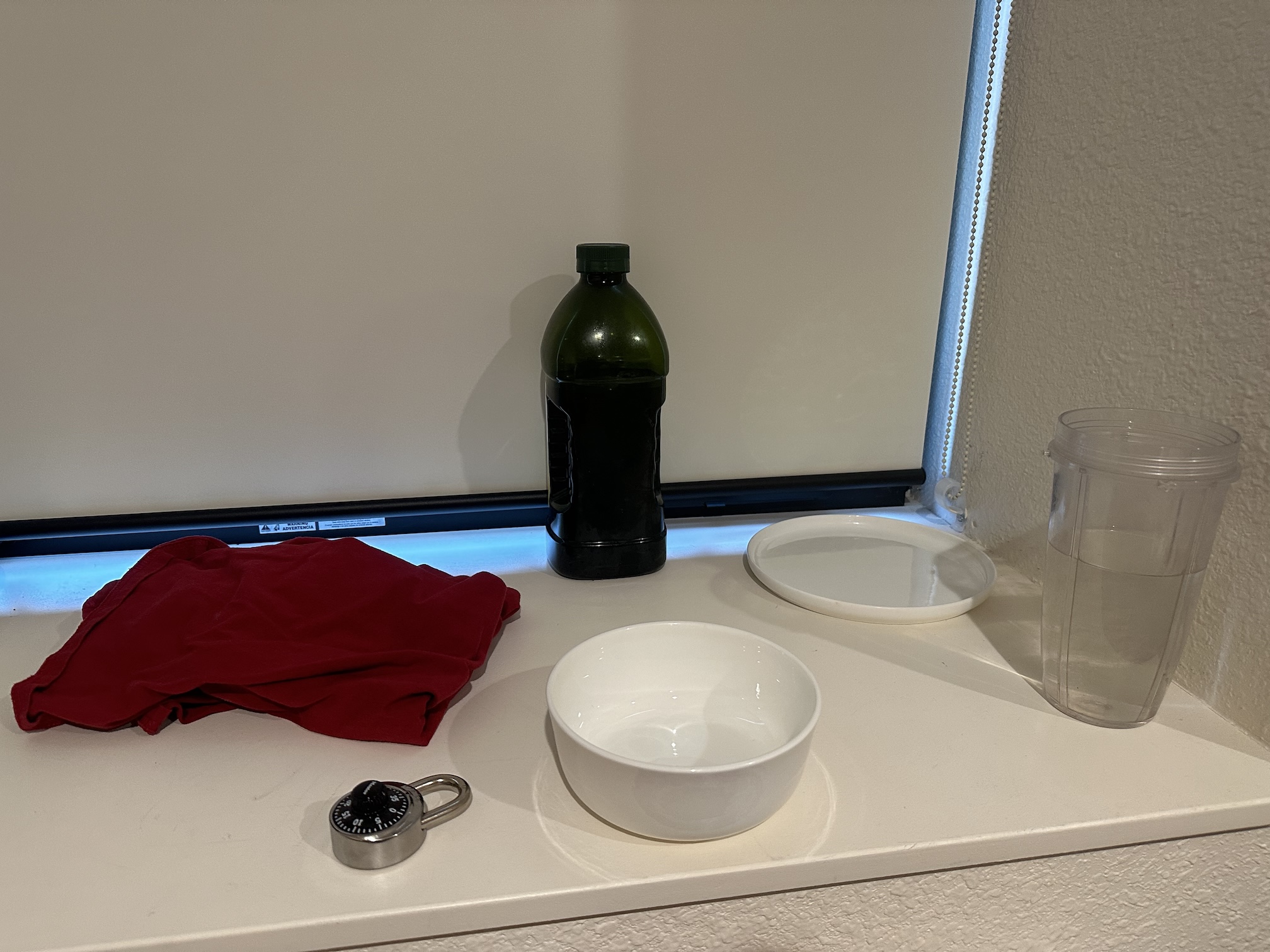}} &
        \begin{enumerate}[leftmargin=*]
            \item bottle
            \item pitcher (container)
            \item bowl [flatter bowl]
            \item towel [shirt]
            \item countertop
            \item bowl [taller ceramic bowl]
            \item measuring cup [lock]
        \end{enumerate} &
        \begin{enumerate}[leftmargin=*]
            \item Bring me the heaviest object. [S]
            \item Bring me the most deformable object. [S]
            \item Bring me the most fragile object. [S]
            \item Bring me all containers that you can confidently determine have water. [M]
            \item Bring me the container with oil. [M]
            \item Among all empty containers, bring me the ones that cannot be used to carry water. [M]
            \item Bring me the metal object. [S]
        \end{enumerate} \\
        \raisebox{-\totalheight}{\includegraphics[width=0.3\textwidth]{figures/planning_scenes/real2.jpg}} &
        \begin{enumerate}[leftmargin=*]
            \item suitcase [blue crate]
            \item stool
            \item hair dryer [mirror]
            \item chair [chair that the mirror is on]
            \item dishwasher [metal cabinet in top right]
            \item chair [blue chair]
            \item bottle [Elmer glue container]
            \item bottle [Mod Podge container]
            \item container [paint thinner container]
            \item desk
            \item mug [mug with paintbrushes]
            \item facial tissue holder [container with glitter]
            \item pencil
        \end{enumerate} &
        \begin{enumerate}[leftmargin=*]
            \item Bring me the heaviest object. [S]
            \item Bring me a metal container. [M]
            \item Bring me a small, empty cup that I can fill with water to clean my paintbrushes. If there are none, tell me that there are no small empty cups. [M]
            \item Bring me the clear container with art supplies. [C]
            \item Bring me the metal object that is reflective. [M]
            \item Bring me paint thinner. [C]
            \item Bring me a wooden object. [S]
        \end{enumerate} \\
        \bottomrule
    \end{tabular}
    \caption{Scene images, object detections, and task instructions for our real scene planning evaluation (scenes 1 and 2). The object category labels given by OWL-ViT are sometimes inaccurate or ambiguous, in which case we provide more precise labels in square brackets. Note that the planner only has access to the original OWL-ViT labels. Tasks are labeled with S, M, or C for \emph{Single Concept}, \emph{Multi-Concept}, or \emph{Common Knowledge}, respectively.}%
    \label{table:plan_scenes}
\end{table*}

\begin{table*}[ht]
    \ContinuedFloat
    \centering
    \begin{tabular}{p{0.3\textwidth}p{0.3\textwidth}p{0.3\textwidth}}
        \toprule
        \textbf{Scene Image} & \textbf{Object Detections} & \textbf{Task Instructions} \\
        \midrule
        \raisebox{-\totalheight}{\includegraphics[width=0.3\textwidth]{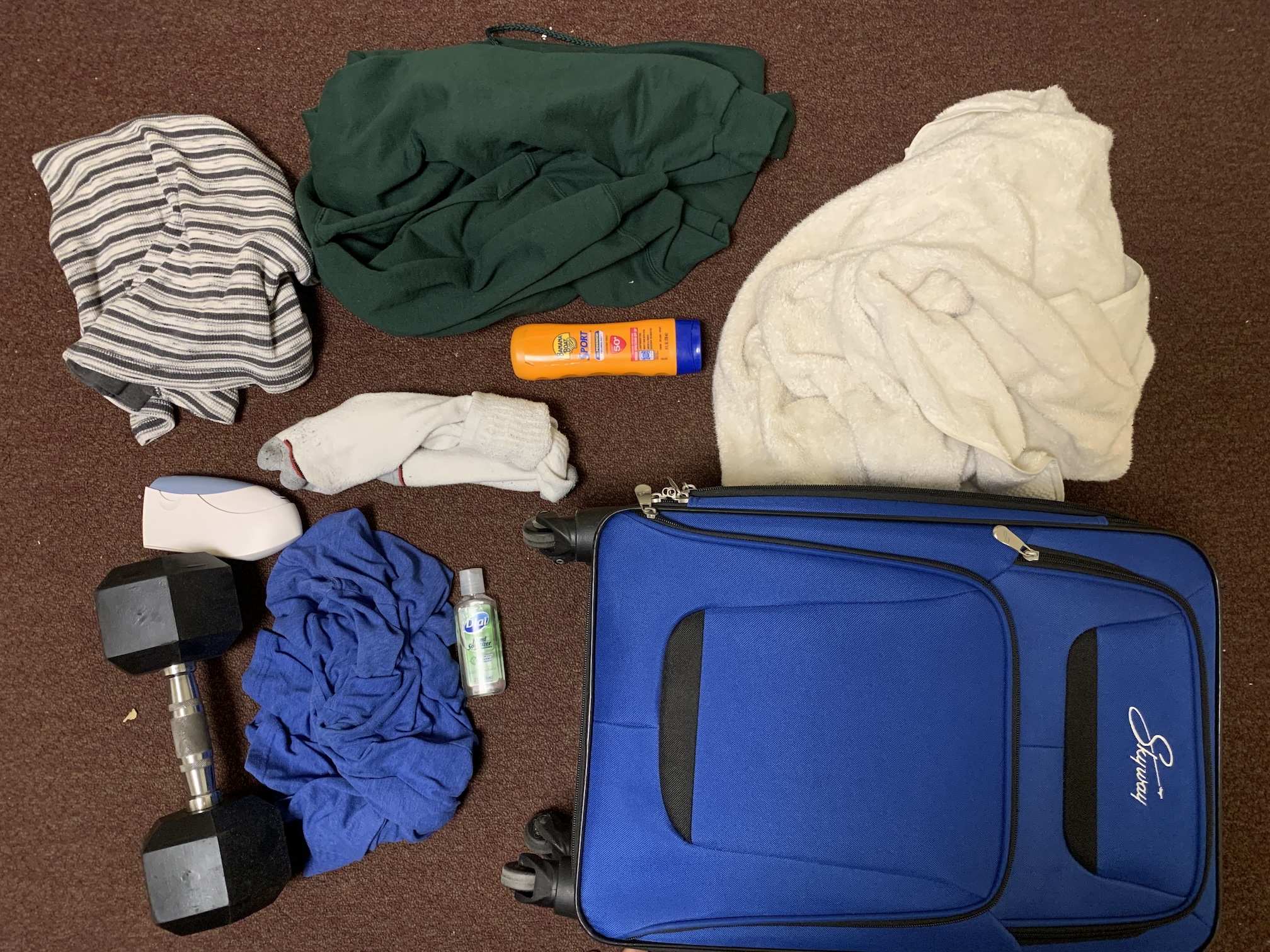}} &
        \begin{enumerate}[leftmargin=*]
            \item clothing [green hoodie]
            \item towel
            \item clothing [striped shirt]
            \item bottle [sunscreen bottle]
            \item towel [socks]
            \item mouse [ear thermometer]
            \item suitcase 
            \item bottle [hand sanitizer]
            \item hair dryer [dumbbell]
            \item clothing [blue shirt]
        \end{enumerate} &
        \begin{enumerate}[leftmargin=*]
            \item Bring me the heaviest object. [S]
            \item Bring me all clear containers. [M]
            \item Bring me the hard plastic object. [M]
            \item Bring me the lightest piece of clothing. [S]
            \item Bring me the object I can pack my clothes into. [C]
            \item It is cold outside. Bring me something that can keep me warm. [C]
            \item It is sunny outside. Bring me the container of sunscreen. [C]
        \end{enumerate} \\
        \raisebox{-\totalheight}{\includegraphics[width=0.3\textwidth]{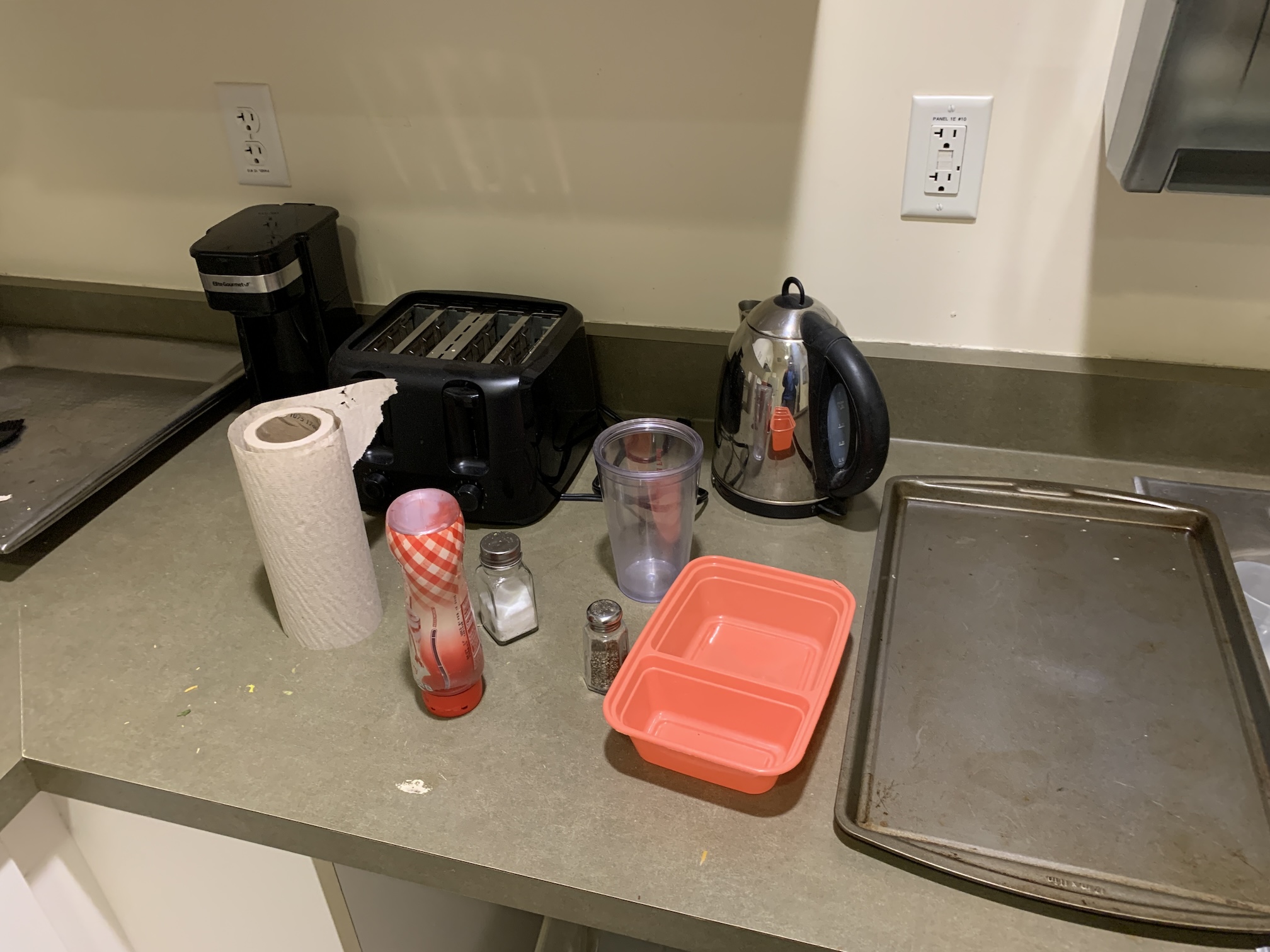}} &
        \begin{enumerate}[leftmargin=*]
            \item facial tissue holder [paper towel dispenser]
            \item light switch [left electric outlet]
            \item light switch [right electric outlet]
            \item mixer
            \item toaster
            \item kettle
            \item paper towel
            \item water glass [plastic cup]
            \item salt and pepper shakers [salt]
            \item bottle [jam container]
            \item frying pan [baking pan]
            \item container [salmon-colored container]
            \item salt and pepper shakers [pepper]
            \item countertop
        \end{enumerate} &
        \begin{enumerate}[leftmargin=*]
            \item Bring me the heaviest object. [S]
            \item Bring me the heavier glass container. [M]
            \item Bring me something that is easy to tear. [C]
            \item Bring me the lightest container that is empty but can be filled with water. [M]
            \item Bring me the most deformable container with a lid. [M]
            \item Bring me all metal containers that can be used to carry water. [M]
            \item Bring me the object that can be used in an oven. [C]
        \end{enumerate} \\
        \bottomrule
    \end{tabular}
    \caption{Scene images, object detections, and task instructions for our real scene planning evaluation (scenes 3 and 4). The object category labels given by OWL-ViT are sometimes inaccurate or ambiguous, in which case we provide more precise labels in square brackets. Note that the planner only has access to the original OWL-ViT labels. Tasks are labeled with S, M, or C for \emph{Single Concept}, \emph{Multi-Concept}, or \emph{Common Knowledge}, respectively.}%
    \label{table:plan_scenes2}
\end{table*}

\begin{table*}[ht]
    \ContinuedFloat
    \centering
    \begin{tabular}{p{0.3\textwidth}p{0.3\textwidth}p{0.3\textwidth}}
        \toprule
        \textbf{Scene Image} & \textbf{Object Detections} & \textbf{Task Instructions} \\
        \midrule
        \raisebox{-\totalheight}{\includegraphics[width=0.3\textwidth]{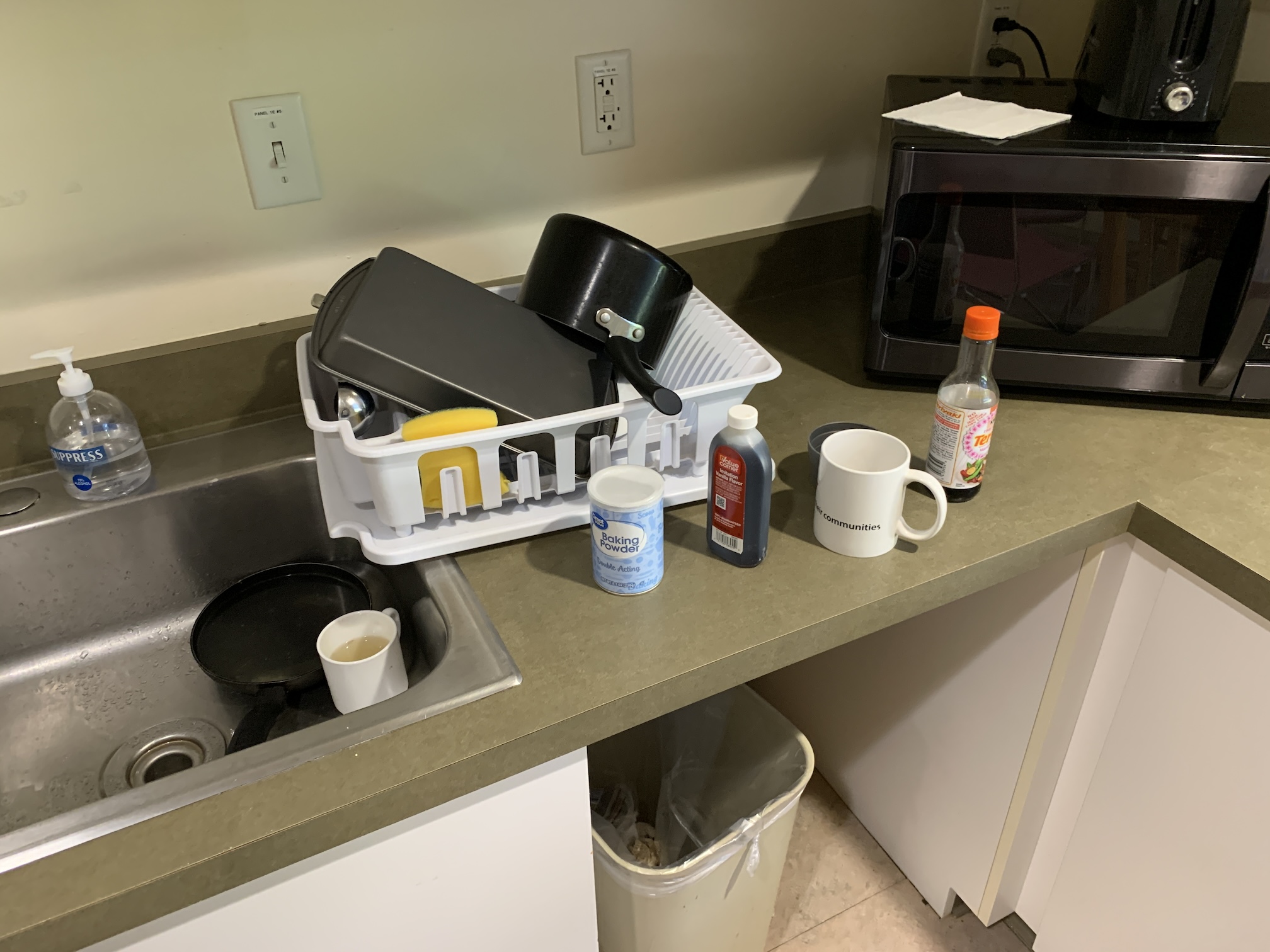}} &
        \begin{enumerate}[leftmargin=*]
            \item toaster
            \item light switch [electric outlet]
            \item envelope [napkin on microwave]
            \item light switch
            \item microwave oven [microwave]
            \item door [microwave door]
            \item bottle [glass sauce bottle]
            \item picnic basket [drying rack]
            \item soap dispenser
            \item bottle [plastic bottle with blue vanilla flavor]
            \item mug [dry mug]
            \item sink
            \item frying pan [dirty pan in sink]
            \item mug [dirty mug in sink]
            \item countertop
            \item waste container
            \item cupboard
            \item plastic bag [trashbag]
        \end{enumerate} &
        \begin{enumerate}[leftmargin=*]
            \item Bring me the heaviest object that you can carry. [S]
            \item Bring me an empty mug that I can use to make tea. [C]
            \item Bring me the most deformable object. [S]
            \item Bring me the glass object. [S]
            \item Bring me a metal pan that is in the sink. [C]
            \item Bring me the container that stores trash. [C]
        \end{enumerate} \\
        \raisebox{-\totalheight}{\includegraphics[width=0.3\textwidth]{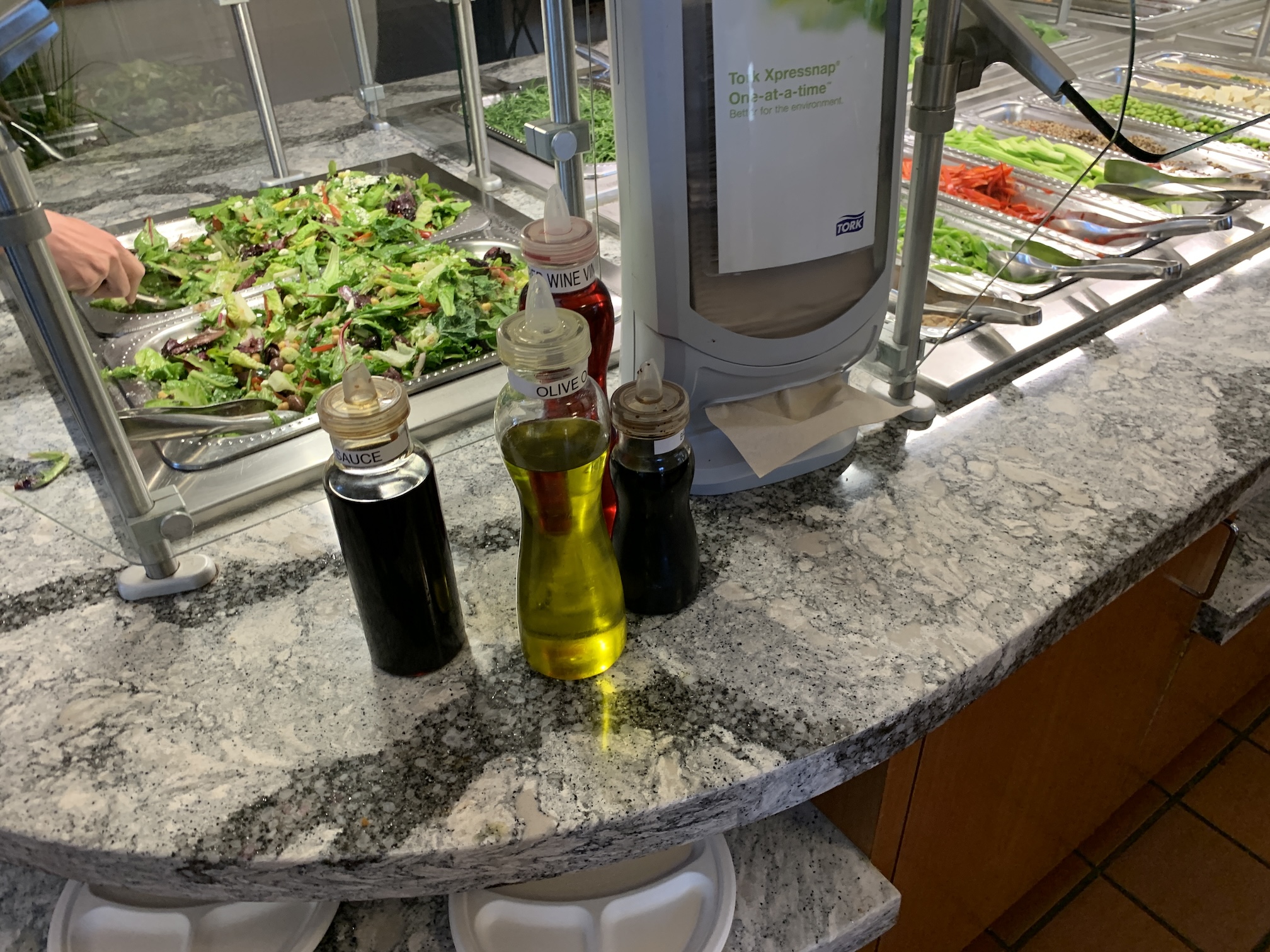}} &
        \begin{enumerate}[leftmargin=*]
            \item envelope [sign on napkin dispenser]
            \item humidifier [napkin dispenser]
            \item ladle [metal tongs]
            \item food [two salad containers on the right]
            \item bottle [red wine vinegar bottle]
            \item frying pan [closer salad tray]
            \item paper [napkin coming out of dispenser]
            \item countertop
            \item bottle [olive oil bottle]
            \item bottle [black container on the right]
            \item bottle [black container on the left]
            \item juice [olive oil inside bottle]
            \item cabinetry
            \item countertop [more cropped in view of countertop]
            \item bowl [paper plate under the counter]
        \end{enumerate} &
        \begin{enumerate}[leftmargin=*]
            \item Bring me the most deformable object. [S]
            \item Bring me the lightest metal object. [M]
            \item Bring me the heaviest glass container. [M]
            \item Serve some food on a plate using objects in the scene. [C]
            \item Bring me an empty container that you can confidently use to contain liquids, if one exists. Otherwise, tell me that no suitable containers exist. [M]
            \item Bring me the container of olive oil. [C]
        \end{enumerate} \\
        \bottomrule
    \end{tabular}
    \caption{Scene images, object detections, and task instructions for our real scene planning evaluation (scenes 5 and 6). The object category labels given by OWL-ViT are sometimes inaccurate or ambiguous, in which case we provide more precise labels in square brackets. Note that the planner only has access to the original OWL-ViT labels. Tasks are labeled with S, M, or C for \emph{Single Concept}, \emph{Multi-Concept}, or \emph{Common Knowledge}, respectively.}%
    \label{table:plan_scenes4}
\end{table*}

\begin{table*}[ht]
    \ContinuedFloat
    \centering
    \begin{tabular}{p{0.3\textwidth}p{0.3\textwidth}p{0.3\textwidth}}
        \toprule
        \textbf{Scene Image} & \textbf{Object Detections} & \textbf{Task Instructions} \\
        \midrule
        \raisebox{-\totalheight}{\includegraphics[width=0.3\textwidth]{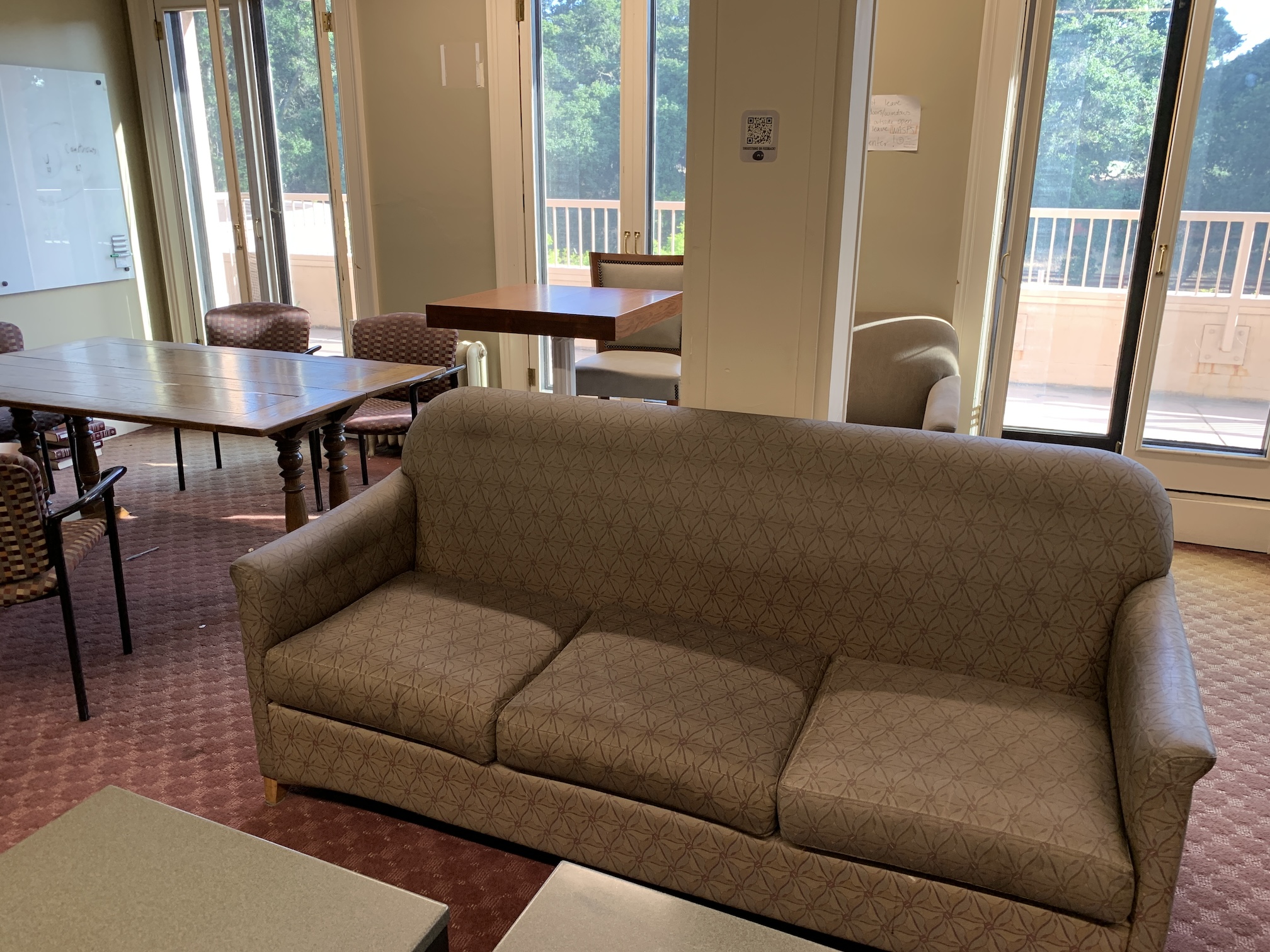}} &
        \begin{enumerate}[leftmargin=*]
            \item whiteboard
            \item door [leftmost door]
            \item paper
            \item window [window of left door of rightmost pair]
            \item door [left door of rightmost pair]
            \item table [taller table]
            \item chair [leftmost short chair facing towards the camera]
            \item chair [tall chair]
            \item chair [short chair behind pillar]
            \item chair [rightmost short chair, facing towards the camera]
            \item table [long wooden table]
            \item door [rightmost door]
            \item couch
            \item chair [left side, facing away from camera]
            \item coffee table
        \end{enumerate} &
        \begin{enumerate}[leftmargin=*]
            \item Go to the piece of furniture that is the softest. [C]
            \item Go to the glass object that is not part of a window or door. [S]
            \item Bring me the lightest object. [S]
            \item Among all pieces of furniture, go to the one that is lightest. [C]
            \item Go to the table that does not have a wooden surface. [S]
        \end{enumerate} \\
        \raisebox{-\totalheight}{\includegraphics[width=0.3\textwidth]{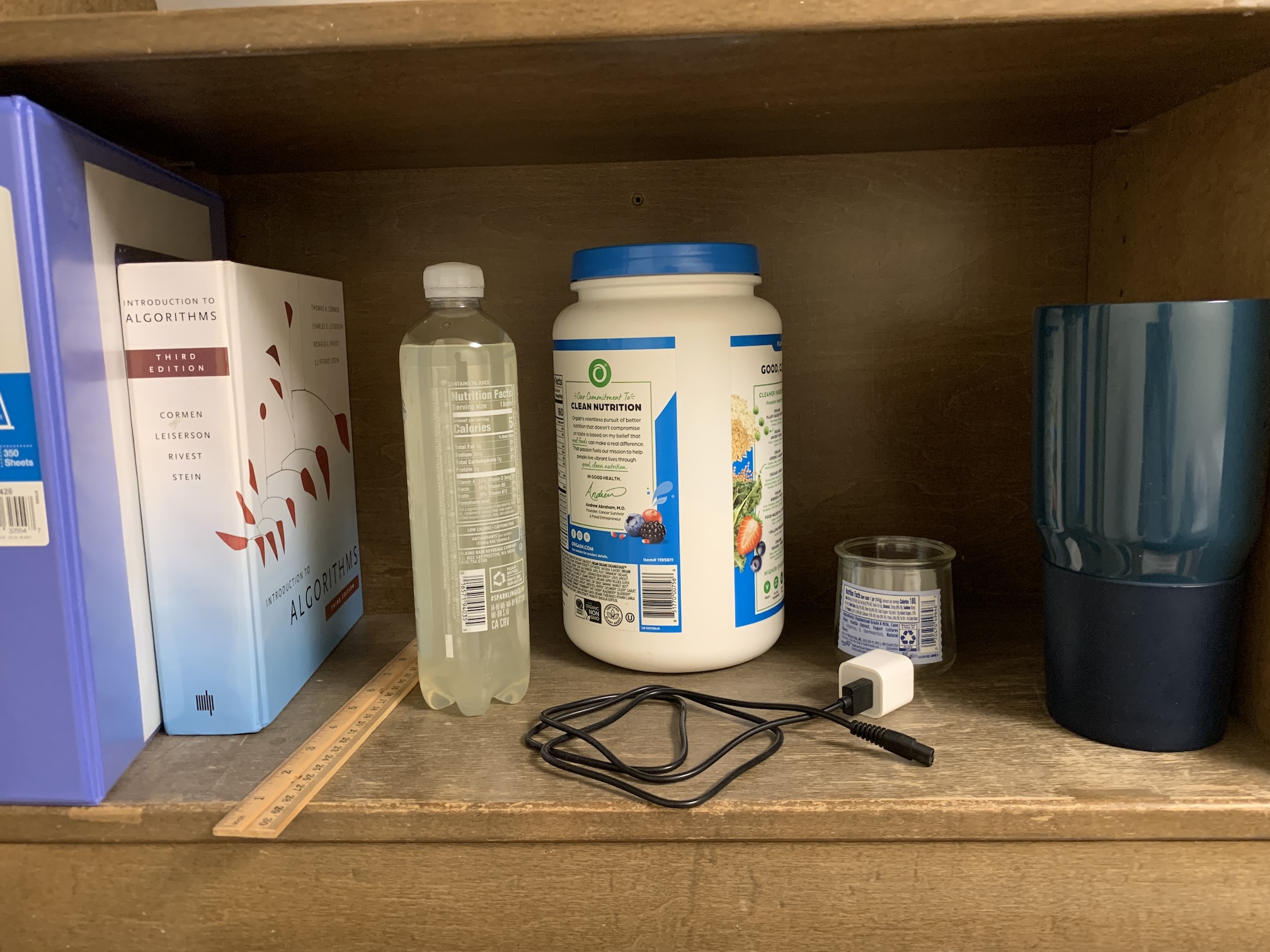}} &
        \begin{enumerate}[leftmargin=*]
            \item box [binder]
            \item bottle [large plastic tub]
            \item bottle [plastic bottle]
            \item box [algorithms textbook]
            \item pitcher (container) [blue metal cup]
            \item water glass [small glass cup]
            \item headphones [phone cable]
            \item dumbbell [power brick]
            \item adhesive tape [ruler]
        \end{enumerate} &
        \begin{enumerate}[leftmargin=*]
            \item Bring me the container that is most likely to be metal. [M]
            \item Bring me the heaviest container. Only consider the container itself, not the contents inside. [S]
            \item Bring me the sealed container with juice. [M]
            \item Bring me the most bendable object. [S]
            \item Bring me the most fragile object. [S]
            \item Bring me all containers that are made of plastic (with very high confidence). [M]
        \end{enumerate} \\
        \bottomrule
    \end{tabular}
    \caption{Scene images, object detections, and task instructions for our real scene planning evaluation (scenes 7 and 8). The object category labels given by OWL-ViT are sometimes inaccurate or ambiguous, in which case we provide more precise labels in square brackets. Note that the planner only has access to the original OWL-ViT labels. Tasks are labeled with S, M, or C for \emph{Single Concept}, \emph{Multi-Concept}, or \emph{Common Knowledge}, respectively.}%
    \label{table:plan_scenes5}
\end{table*}

\onecolumn
\twocolumn
\subsection{Real Robot Evaluation Details}
\label{sec:robot_eval}

\noindent \textbf{Primitives.}
We list the primitives for our real robot evaluation below:
\begin{itemize}
  \item move [X] to the side
  \item move [X] into [Y]
  \item done
\end{itemize}
Similarly as with our planning-only evaluation, our primitives are parameterized by a letter (in place of [X] or [Y]) that identifies each detected object in the scene. The assignment of letters is provided in the list of object detections given to the LLM planner.

\noindent \textbf{Scenes and Tasks.}
In \cref{table:robot_scenes}, we provide the scene images in our real robot evaluation, the detected objects in each scene, and the task instructions for each scene.

\begin{table*}[ht]
    \centering
    \begin{tabular}{p{0.3\textwidth}p{0.31\textwidth}p{0.3\textwidth}}
        \toprule
        \textbf{Scene Image}                                   & \textbf{Object Detections} & \textbf{Task Instructions} \\
        \midrule
        \raisebox{-\totalheight}{\includegraphics[width=0.3\textwidth]{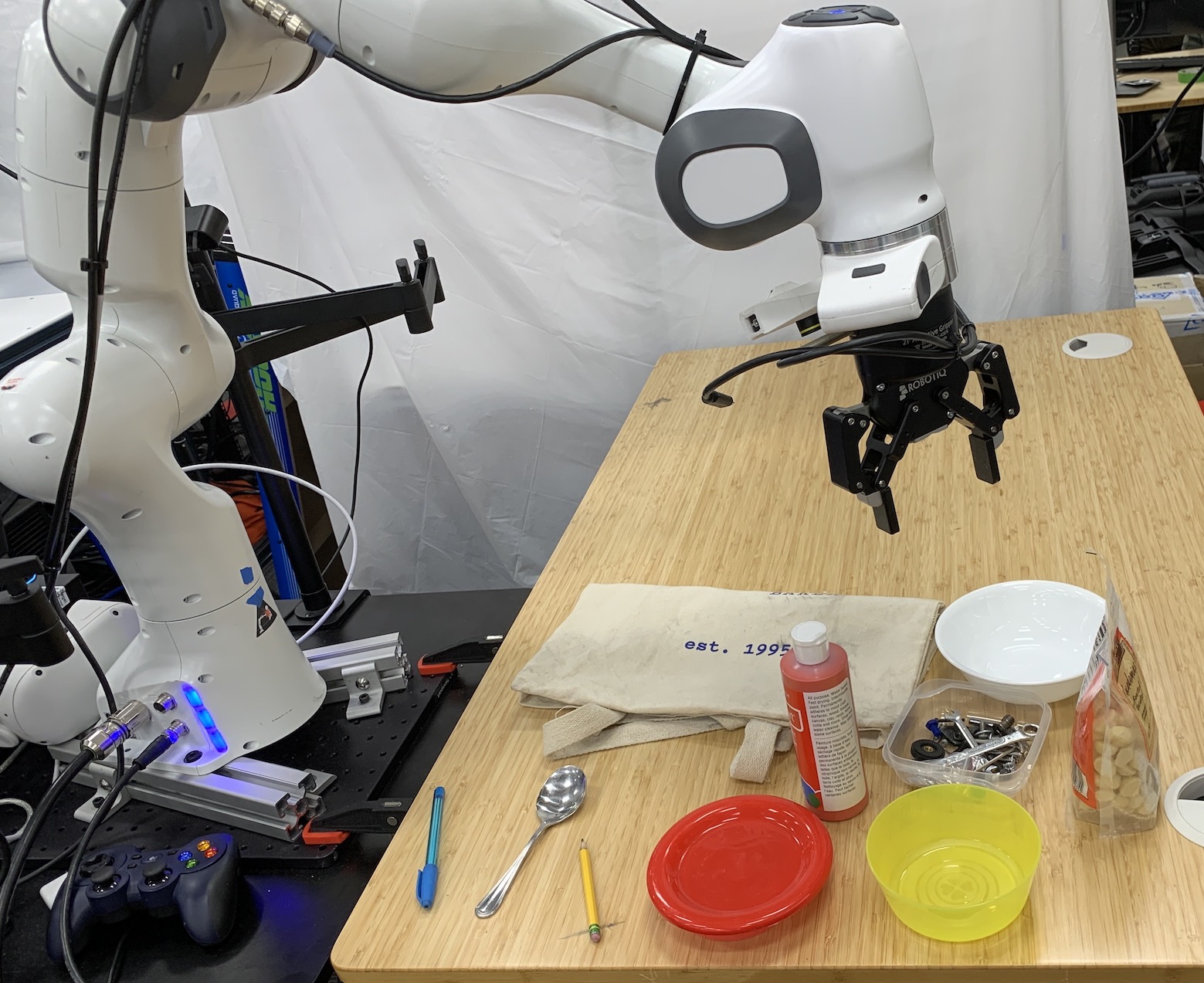}} &
        \begin{enumerate}[leftmargin=*]
            \item towel [handbag]
            \item bottle [paint bottle]
            \item bowl [yellow plastic bowl]
            \item tool [container of metals]
            \item desk [full table]
            \item saucer
            \item bowl [ceramic bowl]
            \item spoon
            \item pencil [pen]
            \item pencil
            \item milk [snack packet]
        \end{enumerate} &
        \begin{enumerate}[leftmargin=*]
            \item Move all objects that are not plastic to the side.
            \item Find a container that has metals. Move all metal objects into that container.
            \item Move all containers that can be used to carry water to the side.
            \item Put the two objects with the least mass into the least deformable container.
            \item Move the most fragile object to the side.
        \end{enumerate} \\
        \raisebox{-\totalheight}{\includegraphics[width=0.3\textwidth]{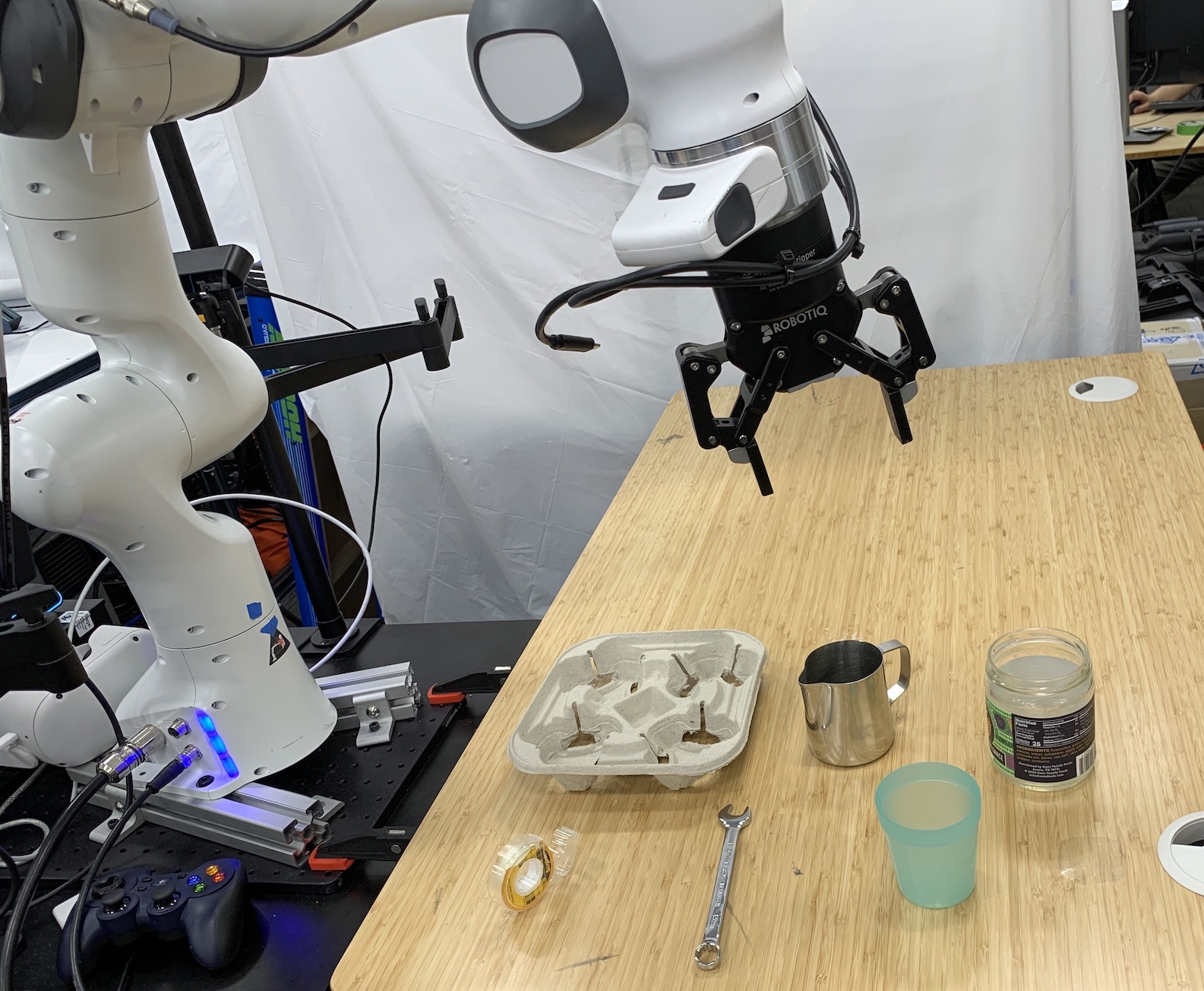}} &
        \begin{enumerate}[leftmargin=*]
            \item bottle [glass jar]
            \item mug [metal mug]
            \item scale [cardboard cupholder]
            \item mug [plastic cup]
            \item adhesive tape
            \item tool [wrench]
        \end{enumerate} &
        \begin{enumerate}[leftmargin=*]
            \item Put all containers that can hold water to the side.
            \item Put all objects that are not plastic to the side.
            \item Put all objects that are translucent to the side.
            \item Put the three heaviest objects to the side.
            \item Put a plastic object that is not a container into a plastic container. Choose the container that you are most certain is plastic.
        \end{enumerate} \\
        \bottomrule
    \end{tabular}
    \caption{Scene images, object detections, and task instructions for our real robot evaluation. The object category labels given by OWL-ViT are sometimes inaccurate or ambiguous, in which case we provide more precise labels in square brackets. Note that the planner only has access to the original OWL-ViT labels.}%
    \label{table:robot_scenes}
\end{table*}

\noindent \textbf{Prompts.}
We use the same prompt structure from the real scene planning evaluation. For tasks that involve moving an object to a side, we add ``In your plan, you may only use the following primitive: move X to the side (where X is an object). Do not move furniture.'' For tasks that involve moving an object into a container, we add ``In your plan, you may only use the following primitive: move X into Y (where X is an object and Y is a container). Do not move furniture.''

\noindent \textbf{Evaluation Procedure.}
We run real robot experiments using a 7-DoF Franka Emika Panda robot with a Robotiq 2F-85 gripper, using Polymetis \cite{Polymetis2021} for real-time control. We obtain our pick-and-place primitives by collecting a kinesthetic demonstration for each primitive, and replaying the demonstration to execute it. The objects in the images provided to the object detector are not in the exact same positions as when the robot is acting, because the objects have to be rearranged when collecting demonstrations for each primitive. However, this does not affect the planner because we do not provide it object positions, as our planning framework does not make use of them.

Because our evaluation focuses on planning quality and not successful execution of primitives, we retry execution of each plan until all of the primitives are successfully executed. Therefore, our success rates are only reflective of planning quality, and not that of the primitives.

We evaluate the success rate of robot executions using a non-author human evaluator. For each evaluation, the evaluator is given the task instruction and a video of the robot executing the generated plan, and they are asked to evaluate whether the robot successfully performed the task instruction. We provide visualizations of robot executions on our \href{\website}{website}.
\end{document}